\documentclass[final]{cvpr}

\usepackage{times}
\usepackage{epsfig}
\usepackage{graphicx}
\usepackage{amsmath}
\usepackage{amssymb}
\usepackage[para]{footmisc}

\usepackage{subfig}
\usepackage{booktabs}
\usepackage{color}
\usepackage{sidecap}
\usepackage{stmaryrd}
\usepackage{floatrow}
\usepackage{enumitem}

\DeclareMathOperator*{\argmin}{arg\,min}

\usepackage[pagebackref=true,breaklinks=true,colorlinks,bookmarks=false]{hyperref}

\def\cvprPaperID{6171} 
\def\confYear{CVPR 2021}

\begin{document}

\title{Temporal-Relational CrossTransformers for Few-Shot Action Recognition}

\author{Toby Perrett
\and Alessandro Masullo \and Tilo Burghardt \and Majid Mirmehdi \and Dima Damen \and 

{\tt\small <first>.<last>@bristol.ac.uk} \quad Department of Computer Science, University of Bristol, UK
}

\maketitle

\begin{abstract}
We propose a novel approach to few-shot action recognition, finding temporally-corresponding frame tuples between the query and videos in the support set. Distinct from previous few-shot works, we construct class prototypes using the CrossTransformer attention mechanism to observe relevant sub-sequences of all support videos, rather than using class averages or single best matches. Video representations are formed from ordered tuples of varying numbers of frames, which allows sub-sequences of actions at different speeds and temporal offsets to be compared.\footnote{Code is available at \url{https://github.com/tobyperrett/TRX}
} 

Our proposed Temporal-Relational CrossTransformers~(TRX) achieve state-of-the-art results on few-shot splits of Kinetics, Something-Something V2 (SSv2), HMDB51 and UCF101.  Importantly, our method outperforms prior work on SSv2 by a wide margin (12\%) due to the its ability to model temporal relations. A detailed ablation showcases the importance of matching to multiple support set videos and learning higher-order relational CrossTransformers.
\end{abstract}

\begin{figure}
\vspace{-2mm}
\centering
\includegraphics[width=0.99\columnwidth]{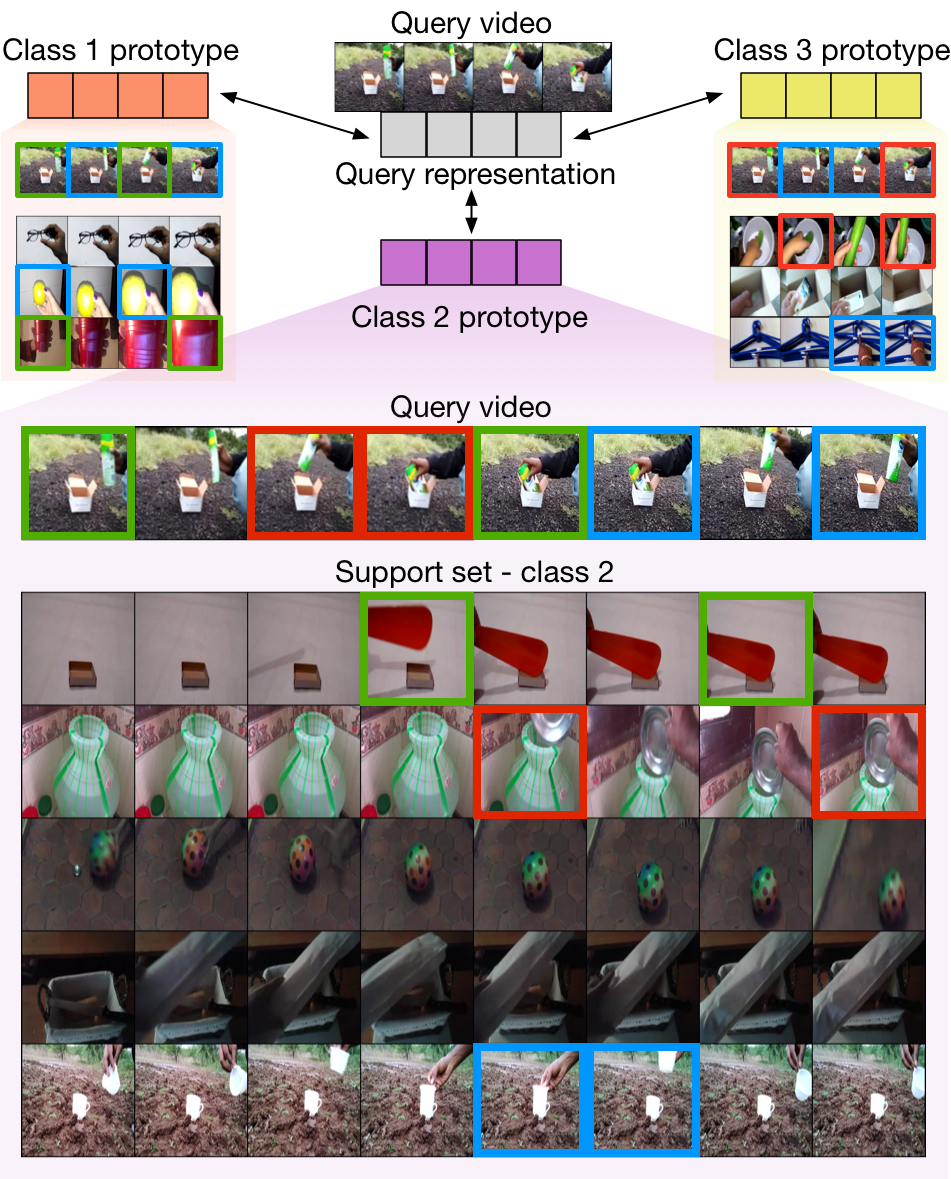}
\caption{For a 3-way 5-shot example, pairs of temporally-ordered frames in the query (red, green, blue) are compared against all pairs in the support set (max attention with corresponding colour). Aggregated evidence is used to construct query-specific class prototypes. We show a correctly-recognised query using our method from SSv2 class ``Failing to put something into something because it does not fit''. \vspace{-8mm}
}
\label{fig:vis}
\end{figure}

\section{Introduction}

Few-shot methods aim to learn new classes with only a handful of labelled examples. 
Success in few-shot approaches for image classification \cite{Finn2017,Requeima2019,Doersch2020} and object recognition \cite{Xu2017,Kang2019} has triggered recent progress in few-shot video action recognition \cite{Zhu2018,Zhu2020,Bishay2019,Zhang2020,Cao2020}. This is of particular interest for fine-grained actions where collecting enough labelled examples proves challenging \cite{Carreira,Goyal2017,Damen2018EPICKITCHENS}.

Recent approaches that achieve state-of-the-art performance~\cite{Bishay2019,Zhang2020,Cao2020} acknowledge the additional challenges in few-shot video recognition, due to varying action lengths and temporal dependencies.
However, these match the query video (\ie the video to be recognised) to the single best video in the support set (\ie the few labelled examples per class), e.g.~\cite{Zhang2020}, or to the average across all support set videos belonging to the same class~\cite{Bishay2019,Cao2020}.  Inspired by part-based few-shot image classification \cite{Doersch2020}, we consider that, within a few-shot regime, it is advantageous to compare sub-sequences of the query video to sub-sequences of all support videos when constructing class prototypes. 
This better accumulates evidence, by matching sub-sequences at various temporal positions and shifts.

We propose a novel approach to few-shot action recognition, which we term Temporal-Relational CrossTransformers~(TRX).  A query-specific class prototype is constructed by 
using an attention mechanism to match each query sub-sequence against all sub-sequences in the support set, and aggregating this evidence.
By performing the attention operation over temporally-ordered sub-sequences rather than individual frames (a concept similar to that in many-shot action-recognition works, e.g.~\cite{Zhou2018, Feichtenhofer2019}), we are better able to match actions performed at different speeds and in different parts of videos, allowing distinction between 
fine-grained classes.  Fig.~\ref{fig:vis} shows an example of how a query video attends to multiple support set videos using temporally-ordered tuples.

\noindent Our key contributions can be summarised as follows:
\vspace{-1mm}
\begin{itemize}[leftmargin=*,itemsep=-3ex,partopsep=1ex,parsep=1ex]
    \setlength\itemsep{-0.15em}
    \item We introduce a novel method, called the Temporal-Relational CrossTransformer (TRX), for few-shot action recognition.
    \item We combine multiple TRXs, each operating over a different number of frames, to exploit higher-ordered temporal relations (pairs, triples and quadruples).
    \item We achieve state-of-the-art results on the few-shot benchmarks for Kinetics \cite{Carreira}, Something-Something V2 (SSv2)~\cite{Goyal2017}, HMDB51 \cite{Kuehnea} and UCF101 \cite{Soomro2012}.
    \item We perform a detailed ablation, demonstrating how TRX utilises multiple videos from the support set, of different lengths and temporal shifts. 
    Results show that using tuple representations improves over single-frames by 5.8\% on SSv2 where temporal ordering proves critical. 

\end{itemize}

\section{Related Work}\label{sec:related}
\vspace{-1pt}

Few-shot classification methods have traditionally fallen into one of three categories - generative \cite{Zhang2018,Dwivedi2019}, adaptation-based \cite{Finn2017,Nichol} and metric-based \cite{Vinyals2016,Snell2017}.  Generative methods use examples from the target task to generate additional task-specific training data with which to fine-tune a network.  Adaptation-based methods (e.g. MAML \cite{Finn2017}) aim to find a network initialisation which can be fine-tuned with little data to an unseen target task.
Metric-based methods (e.g. Prototypical \cite{Snell2017} or Matching \cite{Vinyals2016} Networks) aim to find a fixed feature representation in which target tasks can be embedded and classified. 

Recent works which perform well on \textbf{few-shot image classification} have found that it is preferable to use a combination of metric-based feature extraction/classification combined with task-specific adaptation \cite{Requeima2019, Bateni2020}.
Most relevant to this paper, the recently introduced CrossTransformer~\cite{Doersch2020} uses an attention mechanism to align the query and support set using image patch co-occurrences.  This is used to create query-specific class prototypes before classification within a prototypical network \cite{Snell2017}. Whilst this is effective for few-shot image classification, one potential weakness is that relative spatial information is not encoded. For example, it would not differentiate between a bicycle and a unicycle. 
This distinction is typically not needed in \cite{Doersch2020}'s tested datasets \cite{triantafillou2019metadataset}, where independent part-based matching is sufficient to distinguish between the classes.

\textbf{Few-shot video action recognition} methods have had success with a wide range of approaches, including memory networks of key frame representations~\cite{Zhu2018,Zhu2020} and
adversarial video-level feature generation \cite{Dwivedi2019}.
Recent works have attempted to make use of temporal information. Notably,~\cite{Bishay2019}~aligns variable length query and support videos before calculating the similarity between the query and support set.
\cite{Zhang2020} combines a variety of techniques, including spatial and temporal attention to enrich representations, and jigsaws for self-supervision.  \cite{Cao2020} achieves state-of-the-art performance by calculating query to support-set frame similarities.  They then enforce temporal consistency between a pair of videos by monotonic temporal ordering.  Their method can be thought of as a differentiable generalisation of dynamic time warping.
Note that the above works either search for the single support video \cite{Zhang2020} or average representation of a support class~\cite{Bishay2019,Cao2020} that the query is closest to.
A concurrent work to ours attempts to resolve this through query-centred learning~\cite{Zhu2021}.
Importantly, all prior works perform attention operations on a frame level, as they tend to use single-frame representations.

Compared to all prior few-shot action recognition methods, 
our proposed method attends to all support set videos, using temporal-relational representations from ordered tuples of frames, sampled from the video. By attending to sub-sequences, our method matches actions at different speeds and temporal shifts. Importantly, we use a combination of different CrossTransformers to match tuples of different cardinalities, allowing for higher-order temporal representations. We next describe our method in detail.

\section{Method} \label{sec:method}
\begin{figure*}[t]
\centering
\includegraphics[width=\textwidth]{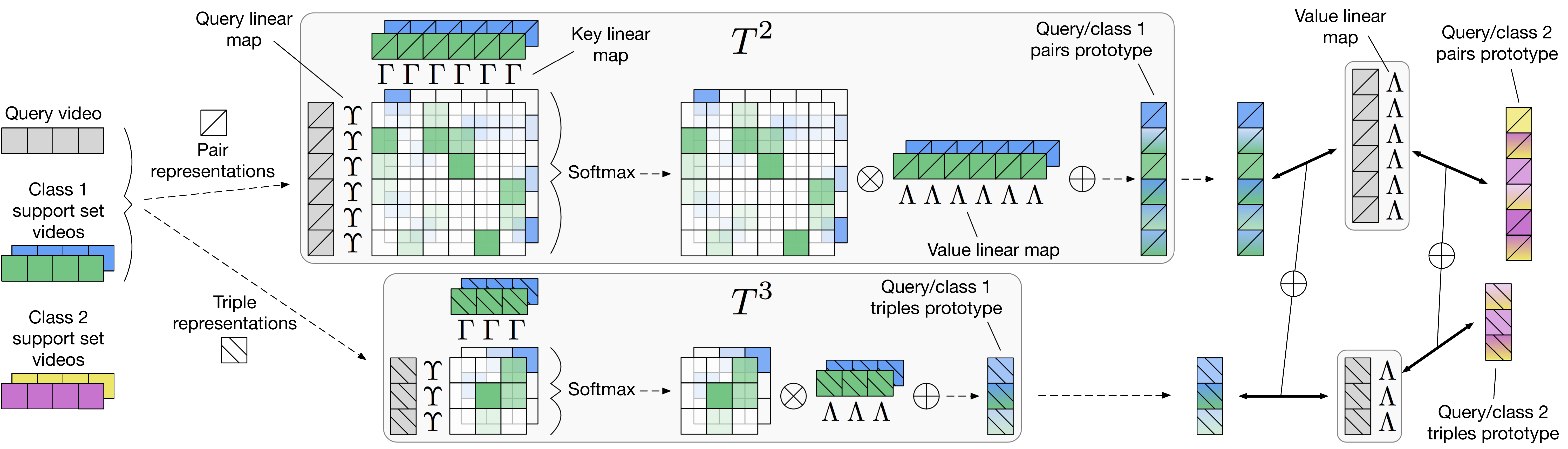}
\caption{Illustration of the Temporal-Relational CrossTransformer (TRX) on a 2-way 2-shot problem. First, pair
$\protect\includegraphics[scale=0.3]{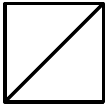}$
and triple  
$\protect\includegraphics[scale=0.3]{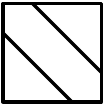}$
representations of the query and support set videos are constructed. 
These are concatenated representations of pairs/triplets of frames (sampled from the video), temporally-ordered.
Two temporal CrossTransformers, $T^2$ and $T^3$, then construct separate class prototypes for each representation (pairs and triplets) using separate query $\Upsilon$, key $\Gamma$ and value $\Lambda$ linear maps for each transformer. This produces a query-specific class prototype, one per CrossTransformer. The query representation is also passed through the value linear maps $\Lambda$, and distances are calculated to the prototypes. The distances are averaged, and the query is recognised as belonging to the closest class. Details are in Section~\ref{sec:method}.\vspace{-8pt}}
\label{fig:method}
\end{figure*}

We propose a method for few-shot action recognition that considers the similarity between an ordered sub-sequence of frames (referred to as a tuple) to all sub-sequences in the support set, through multiple CrossTransformer attention modules. This allows the same query video to match to tuples from several support set videos.
After stating the problem definition in Section \ref{sec:prob_def}, 
for ease of understanding our TRX method, we start from a simplified version, building in complexity and generality up to the full method, which is illustrated in Fig.~\ref{fig:method}.

In Section \ref{sec:single_pair}, we consider a single ordered pair of frames, sampled from the query video.  We propose a temporal CrossTransformer to compare this query pair to ordered pairs of frames from videos in the support set.  This allows the construction of  `query pair'-specific class prototypes.  
We then expand to multiple ordered pairs of frames from the query video. Finally, in Section~\ref{sec:multiple_card},
motivated by the need to model more complex temporal relationships, 
we generalise from pairs to tuples.  We model a separate temporal CrossTransformer for each tuple cardinality to construct query-cardinality-specific class prototypes. These are combined to classify the query video, based on the distances to all class prototypes. 

\subsection{Problem Formulation}\label{sec:prob_def}
In few-shot video classification, inference aims to classify an unlabelled query video into one of several classes, each represented by a few labelled examples unseen in training, referred to as the `support set'. In this paper, we focus on $K$-shot where $K > 1$, i.e. the support set contains more than one video.
Similar to prior works~\cite{Vinyals2016,Finn2017,Cao2020,Zhang2020}, we follow episodic training, i.e. 
random sampling of few-shot tasks from the training set.
For each episode, we consider a $C$-way $K$-shot classification problem.  Let ${Q = \{q_1, \cdots, q_F\}}$ be a query video with $F$ uniformly sampled frames. 
The goal is to classify $Q$ into one of the classes $c \in C$. 
For the class $c$, its support set $S^c$ contains $K$ videos, where the $k^{th}$ video is denoted $S^c_{k} = \{s^c_{k1}, \cdots, s^c_{kF}\}$\footnote{We assume videos are uniformly sampled to be of the same length~$F$ for simplicity.  Alternatively, we could set $F$ to be the maximum video length and non-existent pairs could be masked out in the attention matrix.}. 

\subsection{Temporal CrossTransformer}\label{sec:single_pair}

We consider the temporal relation of two frames sampled from a video to represent the action, as actions are typically changes in appearance and are poorly represented by a single frame.  
We thus sample a pair of ordered frames from the query video with indices $p = (p_1, p_2)$, where ${1 \le p_1 < p_2 \leq F}$, and define the query representation as: 
\begin{equation}\label{eq:single_query}
    Q_p = [\Phi (q_{p_1}) + \text{PE} (p_1), \Phi(q_{p_2}) + \text{PE} (p_2)] \in \mathbb{R}^{2 \times D} ~,
\end{equation}
where $\Phi : \mathbb{R}^{H \times W \times 3} \mapsto \mathbb{R}^D  $ is a convolutional network to obtain a $D$-dimensional embedding of an input frame, and $\text{PE} (\cdot)$ is a positional encoding given a frame index \cite{Vaswani2017}.

We compare the query representation $Q_p$ to all possible pair representations from the support set videos, allowing it to match actions at various speeds and locations within the support set. 
We define the set of all possible pairs as 
\vspace{-3pt}
\begin{equation}\label{eq:single_pair}
    \Pi = \{ (n_1, n_2) \in \mathbb{N}^2 : 1 \leq n_1 < n_2 \leq F)\}.
\vspace{-3pt}
\end{equation}
A single frame-pair representation of video $k$ in the support set of class $c$ with respect to the ordered pair of indices ${m = (m_1, m_2) \in \Pi}$ is 
\vspace{-1pt}
\begin{equation}
S^c_{km} = [\Phi (s^c_{km_1}) + \text{PE} (m_1), \Phi(s^c_{km_2}) + \text{PE} (m_2)] \in \mathbb{R}^{2 \times D}.  
\end{equation}
The set of all pair representations in the support set for class $c$ is
\vspace{-2pt}
\begin{equation}
\mathbf{S}^c = \{ S^c_{km} : (1 \leq k \leq K ) \land (m \in \Pi) \}.
\vspace{-1pt}
\end{equation}

We propose a temporal CrossTransformer $T$, based on the spatial CrossTransformer~\cite{Doersch2020},  but adapted from image patches to frame pairs, to calculate query-specific class prototypes. 
The CrossTransformer includes query $\Upsilon$, key $\Gamma$ and value $\Lambda$ linear maps, which are shared across classes:
\vspace{-2pt}
\begin{equation}\label{eq_heads_pair}
    \Upsilon, \Gamma : \mathbb{R}^{2 \times D} \mapsto \mathbb{R}^{d_k} \hspace{5mm}\text{and}\hspace{5mm}  \Lambda : \mathbb{R}^{2 \times D} \mapsto \mathbb{R}^{d_v} .
\vspace{-2pt}
\end{equation}
The correspondence between the query pair and pair $m$ of support video $k$ in class $c$ is calculated as 
\vspace{-2pt}
\begin{equation}
a^c_{k m p} = L(\Gamma \cdot S^c_{km}) \cdot L(\Upsilon \cdot Q_{p}),
\vspace{-2pt}
\end{equation}
where $L$ is a standard layer normalisation \cite{Ba2016}.
We apply the Softmax operation to acquire the attention map 
\vspace{-5pt}
\begin{equation}\label{eq:softmax}
\tilde a^c_{k m p} = \frac{\exp(a^c_{k m p}) / \sqrt{d_k}}{\sum_{l,n} \exp(a^c_{l n p}) / \sqrt{d_k}}  .
\vspace{-1pt}
\end{equation}
This is then combined with value embeddings of the support set $\mathbf{v}^c_{km}{=}\Lambda \cdot S^c_{km}$, in order to compute the query-specific prototype with respect to the query $Q_p$,  
\vspace{-1pt}
\begin{equation}
\mathbf{t}^c_{p} = \sum_{km} \tilde a^c_{k m p} \mathbf{v}^c_{k m} .
\label{eq:single-tcp}
\vspace{-2pt}
\end{equation}
Now that we have a query-specific class prototype, we calculate the embedding of the query $Q_p$ with the value linear map such that $\mathbf{u}_p{=}\Lambda \cdot Q_{p}$.  This ensures that the query and support representations undergo the same operations.
The CrossTransformer $T$ computes the distance between the query and support set $\mathbf{S}^c$ by passing $Q_p$ such that
\vspace{-1pt}
\begin{equation}\label{eq:t_single}
T(Q_p, \mathbf{S}^c) =  \| \mathbf{t}^c_{p} - \mathbf{u}_{p}  \|   .
\vspace{-1pt}
\end{equation}

Finding a single pair that best represents the action $Q$ is a difficult problem. Instead, we consider multiple pairs of frames from the query video, such that the query representation is defined as $\mathbf{Q} = \{Q_p : p \in \Pi\}$. 
 In Section \ref{sec:ablation_exhaustive}, we compare exhaustive pairs to random pairs of frames.

To calculate the distance between $\mathbf{Q}$ and $\mathbf{S}^c$, we accumulate the distances from all query pairs, i.e.
\begin{equation}\label{eq:multiple}
T(\mathbf{Q}, \mathbf{S}^c) =  \frac{1}{|\Pi|} \sum_{p \in \Pi} T(Q_p, \mathbf{S}^c)  .
\vspace{-1pt}
\end{equation}
During training, negative query-class distances $T$ are passed as logits to a cross-entropy loss.  During inference, the query $Q_p$ is assigned the class of the closest query-specific prototype, i.e. $\argmin_{c} T(Q_p, \mathbf{S^c})$.
Note that the Softmax operation in Eq.~\ref{eq:softmax} is performed separately for each query pair $Q_p$ (\ie matches are scaled separately for each $p$). 
Our Temporal CrossTransformer thus accumulates matches between  pair representations of the query video (hence the term temporal) and all pairs of frames in the support set.  

\subsection{Temporal-Relational CrossTransformers}\label{sec:multiple_card}

A shortcoming of the above method is that an ordered pair of frames might not be the best representation of an action, particularly when fine-grained distinctions between the classes are required.  Consider two classes: ``picking an object up'' vs ``moving an object''.  To discern between these two actions, a method would require at least three frames - i.e. whether the object is put down eventually or remains in hand. Similarly, consider full-body actions such as ``jumping'' vs ``tumbling''.  This highlights the need to explore higher-order temporal representations.

We extend the Temporal CrossTransformer to a Temporal-Relational CrossTransformer (TRX) by considering a sub-sequence of ordered frames of any length.  We use $\omega$ to indicate the length, or cardinality, of a tuple.  For example, $\omega{=}2$ for a pair, $\omega{=}3$ for a triple. We generalise to possible tuples for any $\omega$, such that
\begin{equation}
    \Pi^\omega = \{(n_1, ..., n_\omega)  \in \mathbb{N}^\omega: \forall i (1 \leq n_i < n_{i+1} \leq F) \}
\label{eq:tupleExhaustive}
\end{equation}
The associated query representation with respect to the tuple with indices $p = (p_1, ..., p_\omega) \in {\Pi^\omega}$, generalising the pair representation in Eq.~\ref{eq:single_query}, is
\begin{equation}\label{eq:tuple_rep}
Q_{p}^\omega = [ \Phi(q_{p_1}) + \text{PE}(p_1), ...,  \Phi(q_{p_\omega})  + \text{PE}(p_\omega)] \in \mathbb{R}^{\omega \times D} .
\end{equation}
This is done similarly for the support set representations.

We define the set of cardinalities as  $\Omega$.  For example, pairs, triples and quadruples of frames would be $\Omega{=}\{2, 3, 4\}$.
We use one TRX per cardinality, as parameters can only be defined for a known input dimensionality (e.g. Eq.~\ref{eq_heads_pair}).
Each TRX $T^\omega$
includes query, key and value linear maps corresponding to the dimensionality of $\omega$:
\begin{equation}
     \Upsilon^\omega, \Gamma^\omega : \mathbb{R}^{\omega \times D} \mapsto \mathbb{R}^{d_k} \hspace{3mm}\text{and}\hspace{3mm}  \Lambda^\omega : \mathbb{R}^{\omega \times D} \mapsto \mathbb{R}^{d_v}  .
     \label{eq:omegaDim}
\end{equation}
Each $T ^\omega$ outputs the distance between the query and support set with respect to tuples of cardinality $\omega$.  
We then accumulate distances from the various TRXs, such that:
\begin{equation}
\mathbf{T}^\Omega(Q, \mathbf{S}^c) = \sum_{\omega \in \Omega}  T^\omega  (\mathbf{Q}^\omega, \mathbf{S}^{c\omega}) .
\label{eq:finalLoss}
\end{equation}
Note that averaging the outputs from each TRX first (as in Eq.~\ref{eq:multiple} for $\omega{=}2$) balances the varying number of tuples for each $\omega$.
As with a single TRX, during training, the negative distance for each class is passed as the logit to a cross-entropy loss.  
During inference, the query is assigned the class which is closest to the query with respect to $\mathbf{T}^\Omega$.

\vspace{-10pt}
\paragraph{Summary:}
TRX in its complete form considers a set of cardinalities $\Omega$. For each $\omega \in \Omega$, different linear maps of corresponding dimensions are trained (Eq.~\ref{eq:omegaDim}). These are trained jointly using a single cross-entropy loss, that uses the summed distances (Eq.~\ref{eq:finalLoss}), where the gradient is backpropagated for each~$\omega$~(Eq.~\ref{eq:multiple}).
The gradient is accumulated from each TRX, through the tuple representations, and backpropagated through a convolutional network to update frame representations. TRX is trained end-to-end with shared backbone parameters for all $\omega \in \Omega$, and all tuples.

\section{Experiments}

\subsection{Setup}

\noindent \textbf{Datasets.}
We evaluate our method on four datasets. The first two are Kinetics~\cite{Carreira} and Something-Something V2~(SSv2)~\cite{Goyal2017}, which have been frequently used to evaluate few-shot action recognition in previous works \cite{Zhu2018,Bishay2019,Zhang2020,Cao2020}.  SSv2, in particular, has been shown to require temporal reasoning (e.g.~\cite{Zhou_2018_ECCV,Price,Jiang2019}). We use the few-shot splits for both datasets proposed by the authors of~\cite{Zhu2018,Zhu2020} which are publicly accessible\footnote{https://github.com/ffmpbgrnn/CMN}. In this setup, 100 videos from 100 classes are selected, with 64, 12 and 24 classes used for train/val/test.  We also provide results for the few-shot split of SSv2 used by \cite{Cao2020} which uses 10x more videos per class in the training set.
Additionally, we evaluate our method on HMDB51 \cite{Kuehnea} and UCF101 \cite{Soomro2012}, using splits from \cite{Zhang2020}.

\noindent \textbf{Evaluation.}
TRX particularly benefits from the presence of a number of videos in the support set (\ie few-shot rather than one-shot). We thus evaluate our method on the standard 5-way 5-shot benchmark, and report average results over 10,000 tasks randomly selected from the test sets.
We provide an ablation on X-shot, including one-shot results in the ablation and appendices for completeness.

\noindent \textbf{Baselines.} We give comparisons against four seminal and recent works~\cite{Zhu2018, Bishay2019,Zhang2020,Cao2020}, which reported state-of-the-art in few-shot video action recognition. These have been discussed in Section \ref{sec:related}.

\noindent \textbf{Implementation details.}
We train our TRX model, with all tuple cardinalities and frame-level backbones, end-to-end.
We use a ResNet-50 backbone \cite{He2016} with ImageNet pre-trained weights \cite{Deng2009}, so we are directly comparable to previous methods~\cite{Zhu2018,Zhu2020,Cao2020}\footnote{\cite{Zhang2020} uses Conv-3D features.}. 
We initialise TRX parameters randomly and set $d_k{=}d_v{=}1152$.
The last 2048 dimensional layer from the ResNet forms the frame-level input to the TRX.
These are concatenated into tuples, depending on the length~$\omega$.
Following \cite{Doersch2020}, the query and key linear maps of each transformer share weights, to encourage similarity matching. 
Videos are re-scaled to height 256 and $F{=}8$ frames are sampled uniformly as in \cite{Wanga}.  They are augmented with random horizontal flipping and 224x224 crops. For testing, just a centre crop is used.  

We use SGD with a learning rate of 0.001, training for 10,000 tasks, which takes around 3 hours (apart from the larger SSv2$^*$ split from \cite{Cao2020}, which uses 75,000 tasks). These hyperparameters were determined using the validation set.
We train TRX on four NVidia 2080Ti GPUs.
Due to the number of backbones (e.g. 48 ResNet-50 backbones when considering 5-shot support set, and a query, with 8 frames each),
we can only fit a single task in memory.
We thus average gradients and backpropagate once every 16 iterations.

\subsection{Results}

\begin{table}[t]
\small
\centering 
\resizebox{1\linewidth}{!}{
\begin{tabular}{lccccc}
\toprule
Method                             &  Kinetics  &   SSv2$^\dagger$  &   SSv2$^*$ & HMDB & UCF \\ \midrule
CMN \cite{Zhu2018}      &  78.9    & - & - & - & -    \\ 
CMN-J \cite{Zhu2020}   &  78.9    & 48.8  & - & - & -    \\ 
TARN \cite{Bishay2019}   &  78.5    & - & - & - & -    \\ 
ARN \cite{Zhang2020}    &  82.4    & - & -   &  60.6    & 83.1    \\
OTAM \cite{Cao2020}     &  85.8    & -  & 52.3 & - & - \\
TRX (Ours)                        &  \bf{85.9}    & \bf{59.1} &  \bf{64.6} &  \bf{75.6}    & \bf{96.1}    \\
\bottomrule
\end{tabular}
}
\vspace{-5pt}
\caption{Results on 5-way 5-shot benchmarks of Kinetics~(split from \cite{Zhu2020}), SSv2 ($^\dagger$: split from \cite{Zhu2020}, $^*$: split from~\cite{Cao2020}), HMDB51 and UCF101 (both splits from \cite{Zhang2020}).}
\label{tab:main_results}
\end{table}

\begin{figure*}
\centering
\subfloat[SSv2$^\dagger$: Throwing something in the air and letting it fall. \label{fig:qsstp1}]{\includegraphics[scale=0.175]{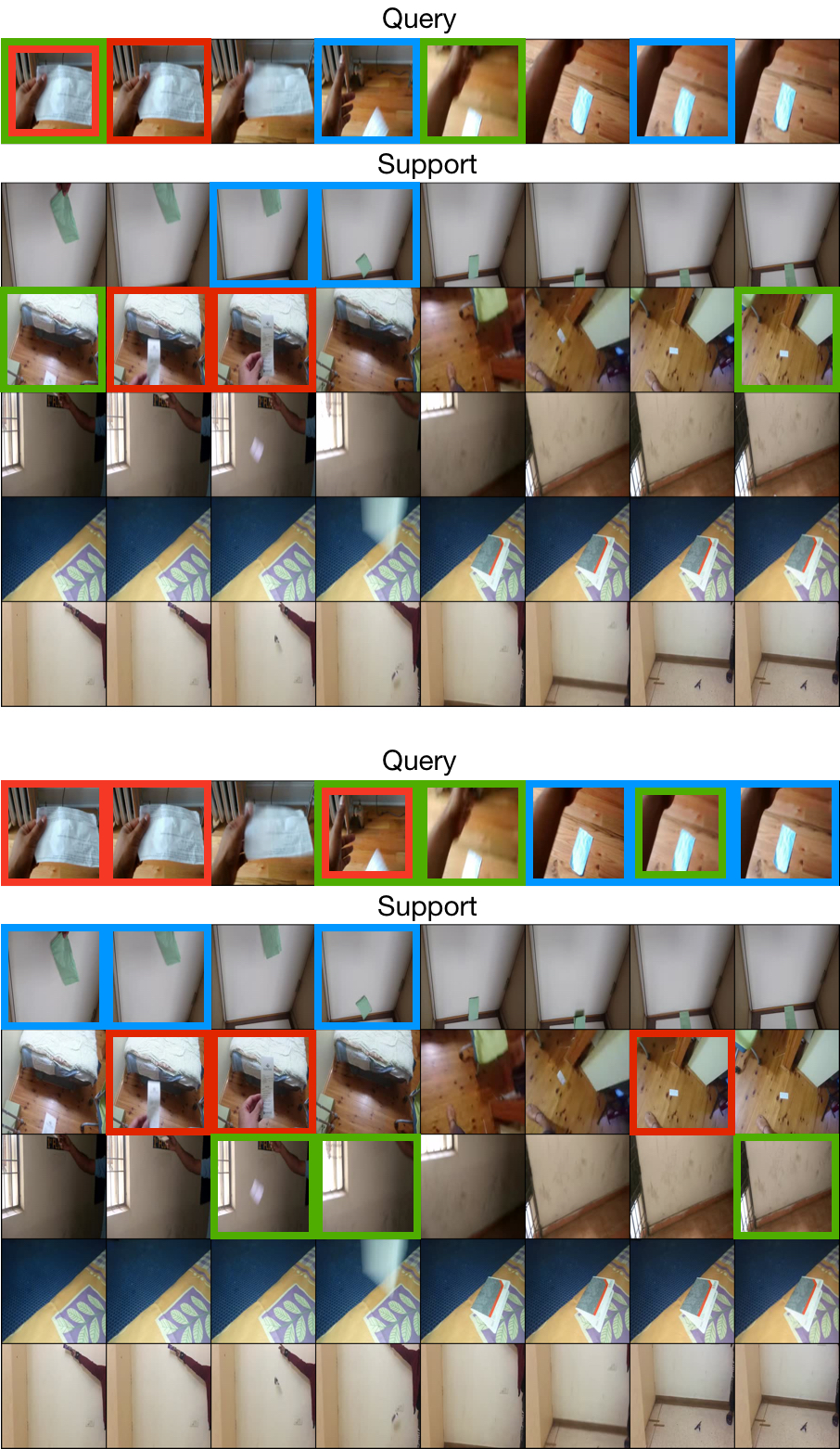}} \hspace{1mm}
\subfloat[Kinetics: Cutting watermelon.\label{fig:qktp1}]{\includegraphics[scale=0.175]{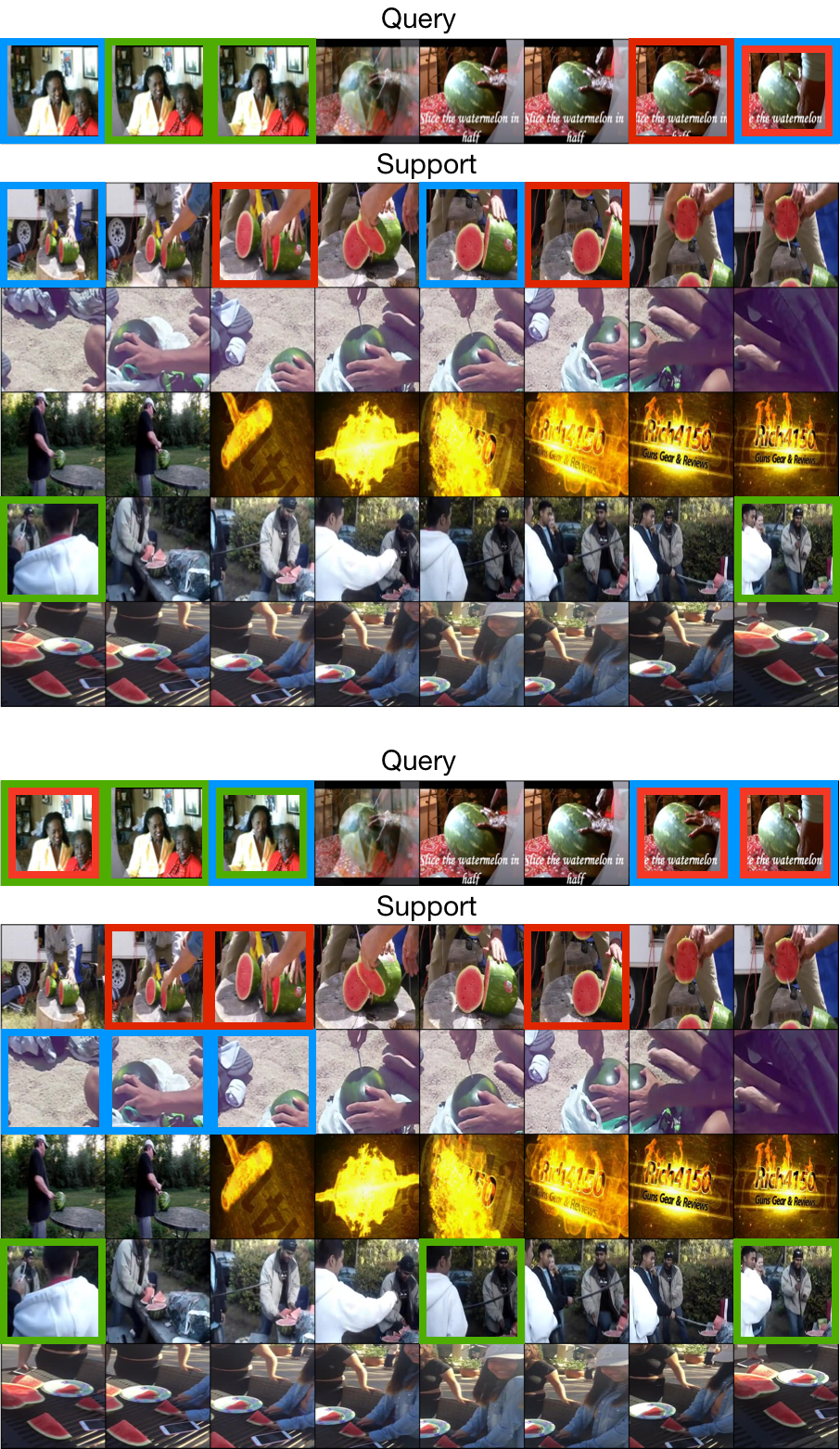}} \hspace{1mm}
\subfloat[SSv2$^\dagger$ (False Positive): Query GT: Failing to put S into S because S does not fit, Support Set: putting something upright on the table.\label{fig:qssfp1}]{\includegraphics[scale=0.175]{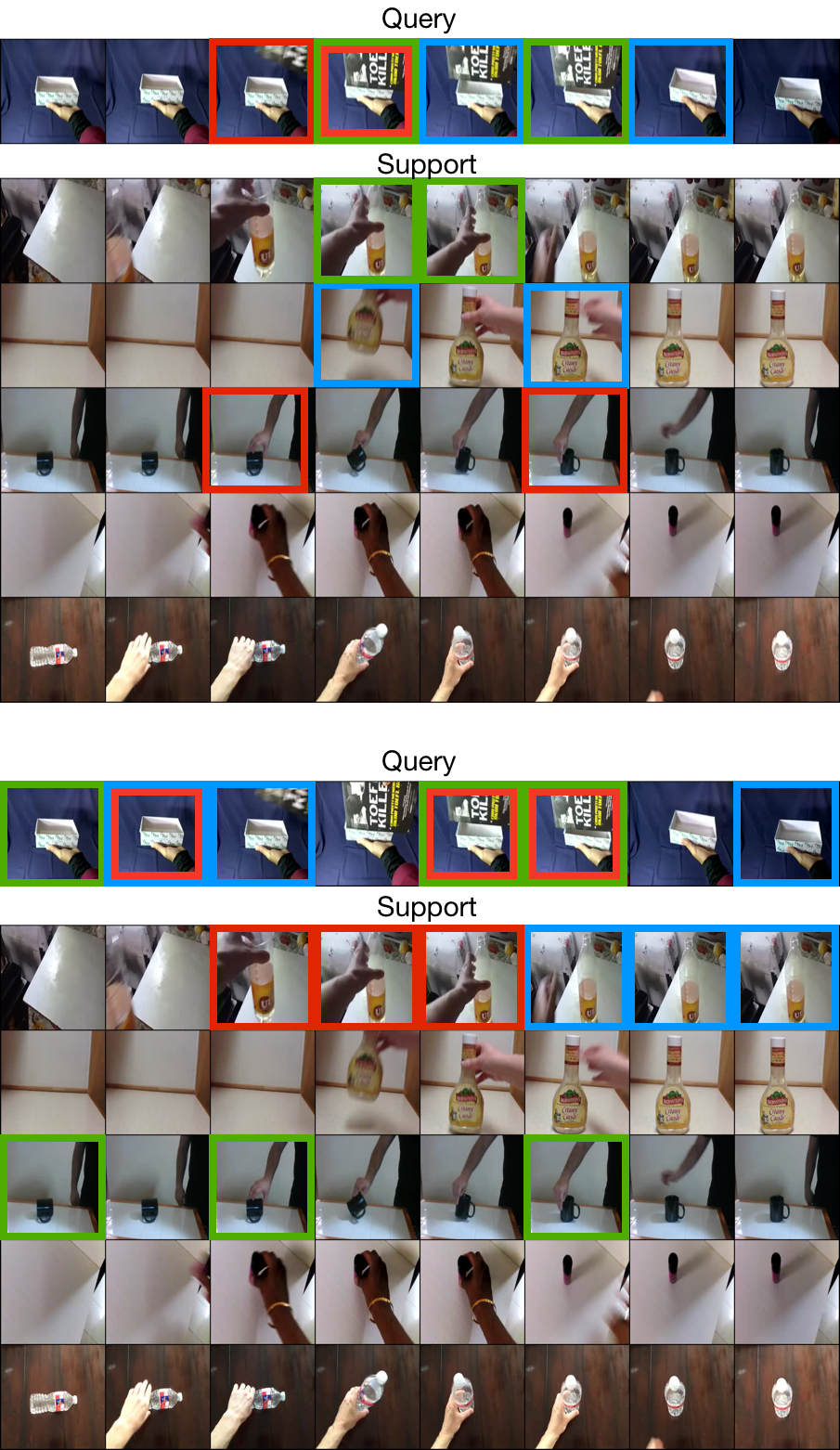}}
\caption{Examples for TRX with $\Omega{=}\{2, 3\}$.  Colour-matching pairs (top) and triplets (bottom) are shown between the query and support set videos from one class.  Three tuples are highlighted in each subfigure (red, green and blue). This figure demonstrates maximum attention matches to several videos in the support set, at different relative and absolute positions.}
    \label{fig:qual}
\end{figure*}

Table~\ref{tab:main_results} shows our comparative results. 
TRX outperforms prior work on the four datasets. On the most challenging dataset (SSv2), TRX outperforms prior work by a wide margin (12\% and 10\% on different splits). 
The large improvement is found on SSv2 because TRX is particularly beneficial when temporally ordered tuples can assist the discrimination between classes. 
It also outperforms prior work on HMDB51 and UCF101 by 15\% and 13\% respectively.
On Kinetics, it exceeds the state-of-the-art (by 0.1\%).  
Kinetics is more of an appearance-based dataset when used as a few-shot benchmark, where ordering is less important and single frame representation can be sufficient.  We ablate this in Section~\ref{sec:ablation_order}.

Figure~\ref{fig:qual} shows qualitative results, highlighting tuple matches between the query and support set for $\Omega{=}\{2, 3\}$. 
For each subfigure, we show query pairs (top) and triplets~(bottom)
with their corresponding tuples (same colour) in the support set. For example, the red pair of frames in the first example (frames 1 and 2) gets the maximum attention when compared to the second support set video (frames 2 and 3). We select three tuples to highlight in each case.
The figure shows that tuples match to different videos in the support set, as well as tuples of varying positions and frame differences.
A failure case (Fig.~\ref{fig:qssfp1}) matches pair/triplet frames from the query ``failing to put something into something because it doesn't fit'', with pairs/triplets of the support set class ``put something upright on the table''. In each example, the \textit{putting} action is correctly matched, but the query is closest to the wrong prototype.

\subsection{Ablations}\label{sec:ablation}

Our motivation in proposing TRX is the importance of representing both the query and the support set videos by tuples of ordered frames, and that class prototypes should be constructed from multiple support set videos. 
We showcase this motivation experimentally through several ablations. We specifically evaluate: (\ref{sec:ablation_cardinalities})~the impact of~$\Omega$, (\ref{sec:ablation_order})~the importance of ordered frames in the tuple, (\ref{sec:ablation_look_vids})~the importance of multiple videos in the support set, and (\ref{sec:ablation_look_pairs})~whether tuples at various locations and frame positions are being matched within TRX. 
Additionally, (\ref{sec:ablation_runtime})~we compare performance and runtime as the number of sampled frames changes, and  (\ref{sec:ablation_exhaustive})~ compare exhaustive to random tuples, showcasing the potential to compress TRX models without a significant drop in performance. We primarily use the large-scale datasets Kinetics and SSv2$^\dagger$ for these ablations using the splits from \cite{Zhu2018,Zhu2020}.
\subsubsection{TRX with different $\Omega$ values}\label{sec:ablation_cardinalities}

\begin{table}
\small
\centering 
\begin{tabular}{lccc}
\toprule
Cardinalities           &   Num tuples  &   Kinetics        &   SSv2$^\dagger$ \\ \midrule
$\Omega{=}\{1\}$        &   -               & 85.2              & 53.3      \\   
$\Omega{=}\{2\}$        &   28              & 85.0              & 57.8          \\ 
$\Omega{=}\{3\}$        &   56              & 85.6              & 58.8          \\ 
$\Omega{=}\{4\}$        &   70              & 84.5              & 58.9          \\ 
$\Omega{=}\{2,3\}$      &   84              &   \textbf{85.9}            & \textbf{59.1}       \\
$\Omega{=}\{2,4\}$      &   98             & 84.4              & 58.4       \\
$\Omega{=}\{3,4\}$      &   126             &  85.3             & \textbf{59.1}       \\
$\Omega{=}\{2,3,4\}$    &   154             &  85.3             & 58.9       \\
\bottomrule
\end{tabular}
\caption{Comparing all values of $\Omega$ for TRX, noting the number of tuples for each model, given by $\sum_{\omega \in \Omega}|\Pi^\omega|$.}
\vspace*{-5pt}
\label{tab:cardinalities_1}
\end{table}

In our comparative analysis in Tab.~\ref{tab:main_results}, we reported results using $\Omega{=}\{2,3\}$. This is the combined TRX of pair and triplet frames demonstrated in Fig.~\ref{fig:method}. We now evaluate each cardinality of $\Omega \in \{1,2,3,4\}$ independently as well as all their combinations on both datasets.

In Tab.~\ref{tab:cardinalities_1}, results demonstrate significant improvement in SSv2 moving from single frame comparisons to pair comparisons, of ~(+4.5\%). The performance increases further for triplets (+1.0\%) and only marginally again for quadruples (+0.1\%). 
Combining two CrossTransformers $\Omega{=}\{2,3\}$ performs best.
Using all cardinalities $\Omega{=}\{2,3,4\}$ results in a slight drop in performance (-0.2\%).
Comparatively, differences are smaller on Kinetics, and moving to quadruples drops the performance significantly (-1.4\%) compared to the best TRX combination $\Omega{=}\{2,3\}$.

\begin{figure}[t]
\centering
\includegraphics[scale=0.37, trim={0 30 0 0}]{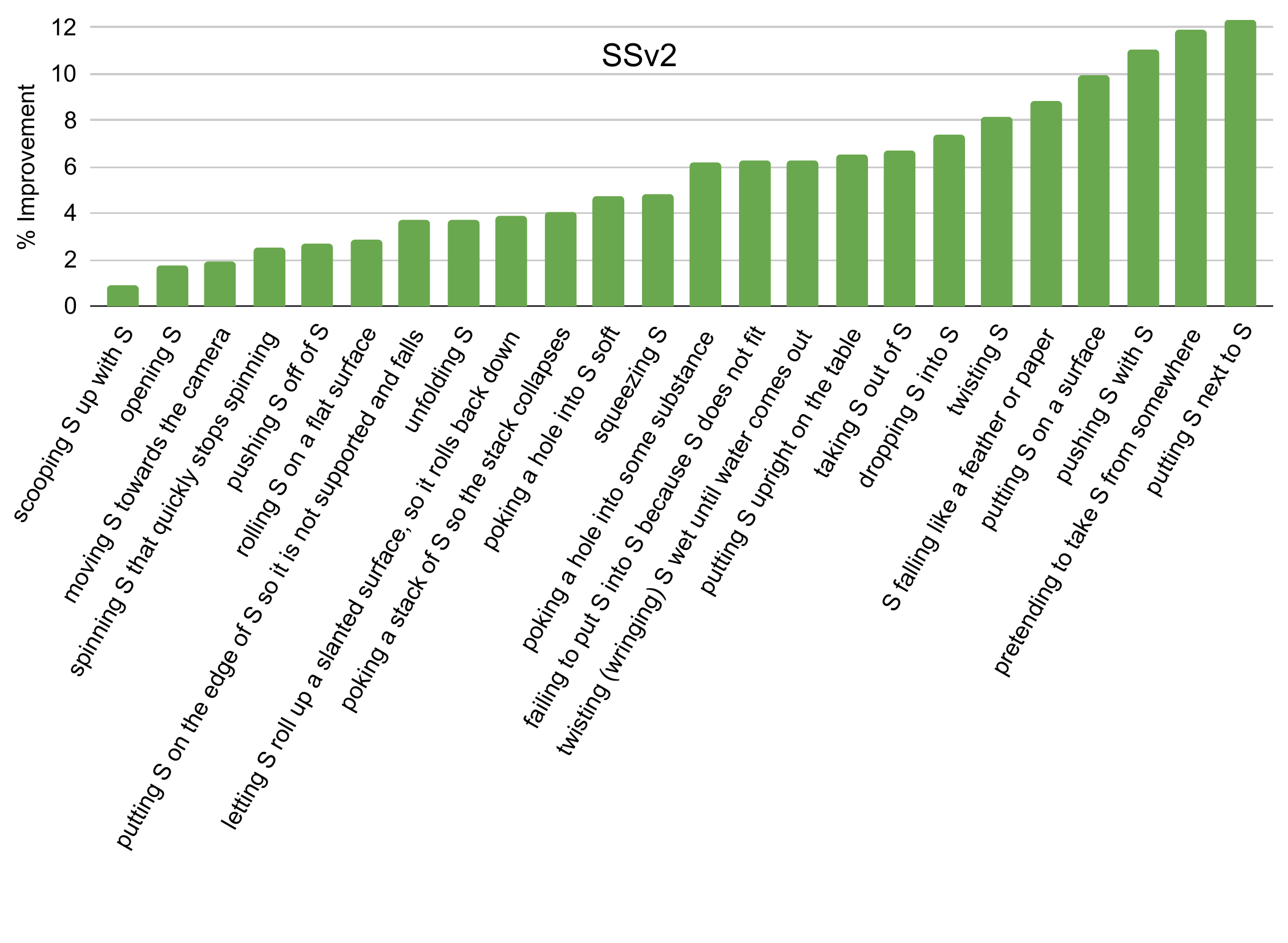}\\
\vspace{-2mm}
\includegraphics[scale=0.37, trim={0 30 0 0}]{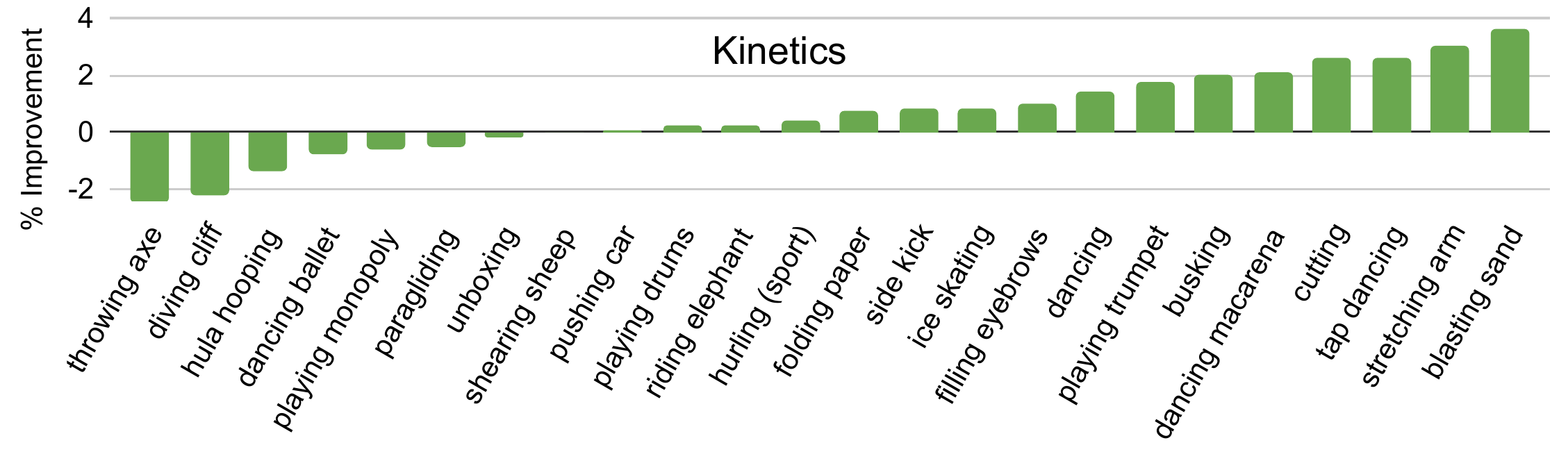}
\caption{Class improvement using tuples ($\Omega{=}\{2,3\}$) compared to single frames ($\Omega{=}\{1\}$) for SSv2$^\dagger$ and Kinetics.\vspace{-8pt}}
\label{fig:class_improvement_ss}
\end{figure}

The improvement using TRX with the multiple cardinalities $\Omega{=}\{2,3\}$ over frame-based comparisons ($\Omega{=}\{1\}$) is demonstrated per-class in Fig.~\ref{fig:class_improvement_ss}.
For SSv2, some classes see little improvement (e.g. ``scooping something up with something'', ``opening something''), whereas others see a greater than 10\% improvement (e.g. ``pretending to take something from somewhere'', ``putting something next to something''). Aligned with the overall results on Kinetics, Fig.~\ref{fig:class_improvement_ss} shows modest improvements per-class, including marginal drop in some classes.

\vspace*{-8pt}
\subsubsection{The impact of ordered tuples}\label{sec:ablation_order}
\vspace*{-5pt}

\begin{table}
\small
\centering 
\begin{tabular}{lcc}
\toprule
Method                                            &   Kinetics    &   SSv2$^\dagger$ \\ \midrule
$\Omega{=}\{2,3\}$ order reversed           &   {\bf 85.9}        & 51.3       \\
$\Omega{=}\{2,3\}$                                &  {\bf 85.9}        & {\bf 59.1}       \\
\bottomrule
\end{tabular}
\vspace{-5pt}
\caption{Results assess the importance of temporal ordering.  When the tuple orders are reversed for the query video, a large drop is observed for SSv2$^\dagger$, but not for Kinetics.\vspace{-14pt}}
\label{tab:ordering}
\end{table}

Up to this point, we have made the assumption that tuples should be temporally ordered to best represent actions.  
We evaluate the extreme scenario, where frames in the support set are temporally ordered, but frames in the query take the reverse order during inference only. 
Table~\ref{tab:ordering} shows a large drop for the reversed query sets on SSv2 (-7.8\%). Supporting our prior observation that it is more an appearance-based dataset, no drop is observed for Kinetics.  

\begin{figure}[t]
\centering
\includegraphics[scale=0.5, trim={0 10 20 5}]{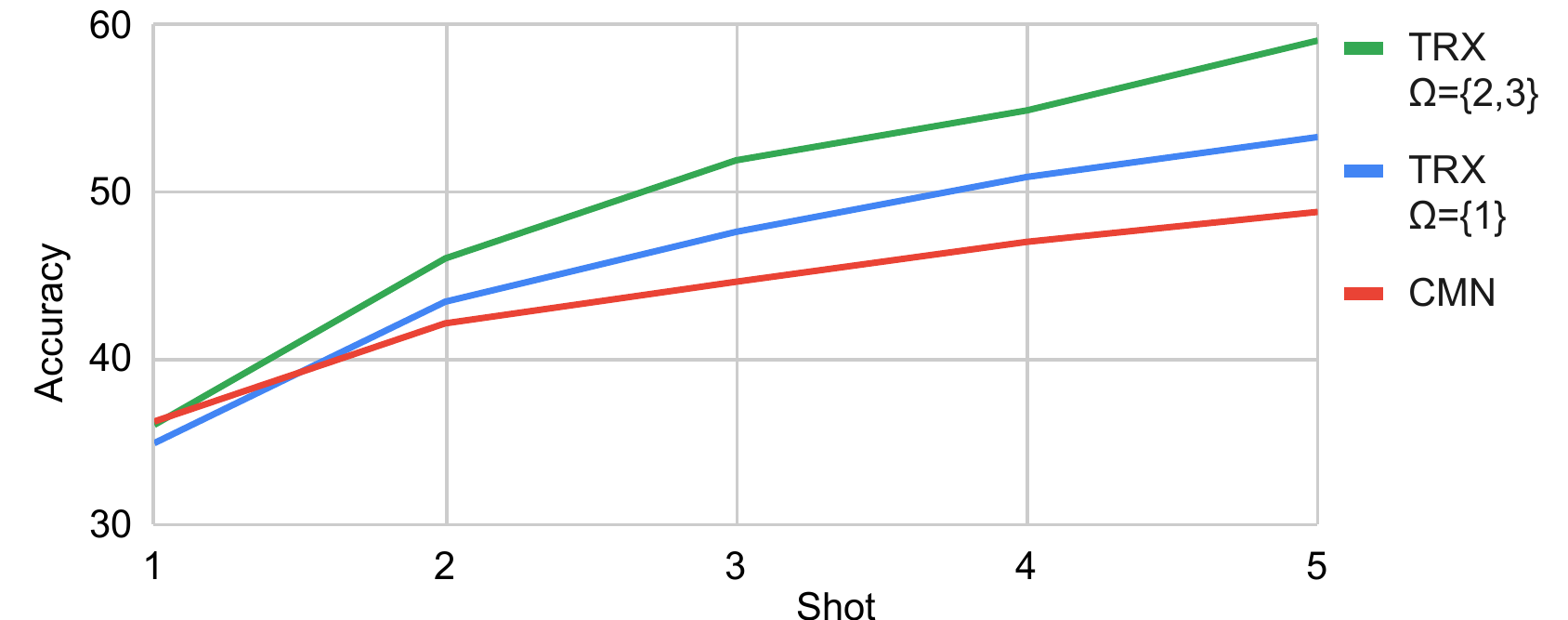}
\caption{Comparing CMN~\cite{Zhu2020} results to TRX for X-shot 5-way, for $1 \le X \le 5$ on SSv2$^\dagger$. TRX clearly benefits from increasing the number of of videos in the support set, both for $\Omega{=}\{1\}$ and using two CrossTransformers $\Omega{=}\{2,3\}$.\vspace*{-14pt}}
\label{fig:kshot}
\end{figure}

\begin{figure}
\centering
\subfloat[SSv2$^\dagger$, $\Omega{=}\{2\}$.\label{fig:ob_ss_2}]{\includegraphics[scale=0.42, trim={0 0 0 20}]{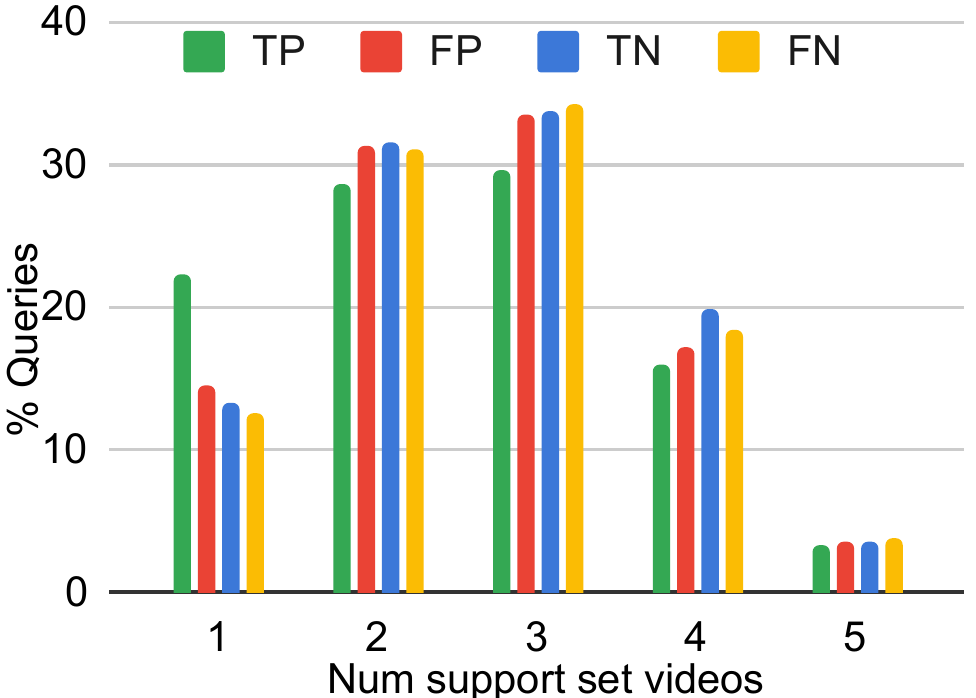}}
\subfloat[SSv2$^\dagger$, $\Omega{=}\{4\}$.\label{fig:ob_ss_4}]{\includegraphics[scale=0.42, trim={0 0 0 20}]{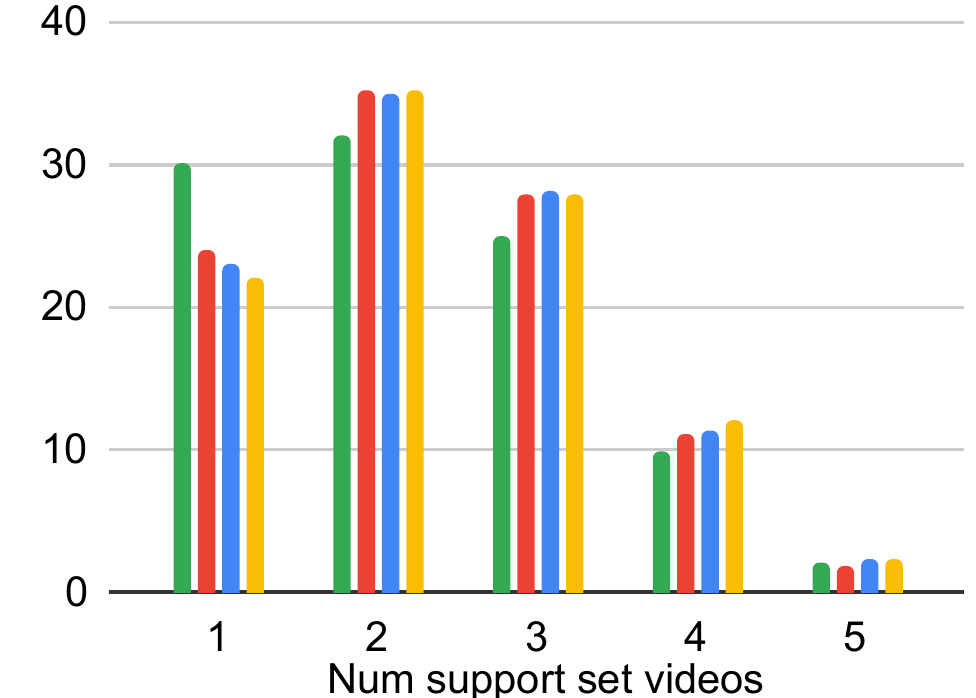}} \\
\subfloat[Kinetics, $\Omega{=}\{2\}$.\label{fig:ob_k_2}]{\includegraphics[scale=0.42, trim={0 0 0 20}]{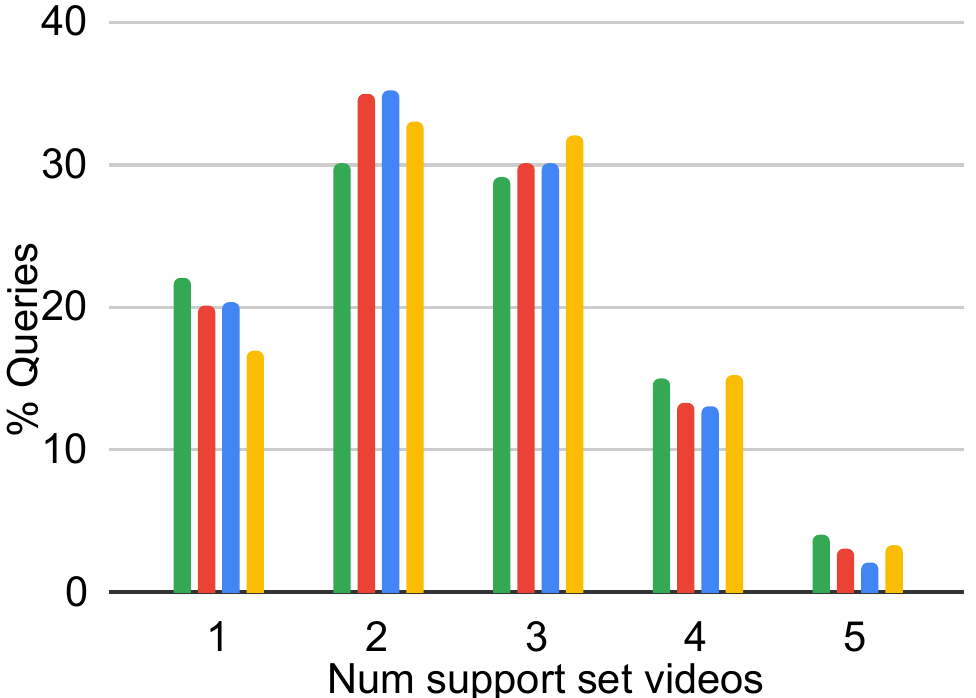}}
\subfloat[Kinetics, $\Omega{=}\{4\}$.\label{fig:ob_k_4}]{\includegraphics[scale=0.42, trim={0 0 0 20}]{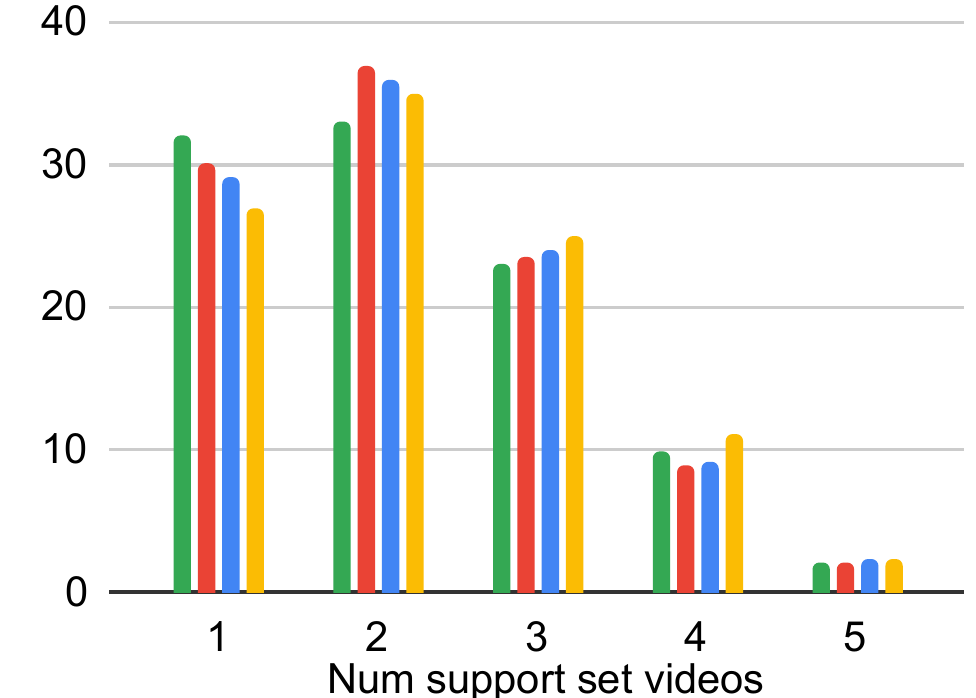}}
\vspace{-5pt}
\caption{Percentage of queries that match a given number of support videos per class, with a max attention value, for SSv2$^\dagger$ and Kinetics.  True/False Positive/Negative query percentages are shown for $\Omega{=}\{2\}$ and $\Omega{=}\{4\}$.}
    \label{fig:observed}
\end{figure}

\subsubsection{Matching to multiple support set videos}\label{sec:ablation_look_vids}
\vspace*{-5pt}

Our motivation for using CrossTransformers is that query tuples would match to tuples from multiple support set videos in order to create the query-specific class prototype. Note that this is not regularised during training - \ie there is no encouragement to use more than one support set video.

Figure~\ref{fig:kshot} ablates TRX ($\Omega{=}\{1\}$ and $\Omega{=}\{2,3\}$) for the number of videos in the support set per class. We increase this from 1-shot to 5-shot reporting the performance for each on SSv2, as well as comparative results from CMN~\cite{Zhu2020}.  Whilst all methods perform similarly for 1-shot, TRX significantly increases the margin over the CMN baseline as the number of shots increases. For our proposed model $\Omega{=}\{2,3\}$, we report improvements of +3.9\%, +7.3\%, +7.9\% and +10.3\% for 2-, 3-, 4- and 5-shots comparatively. Note that using a single frame representation also improves over CMN, by a smaller but significant margin. This ablation showcases TRX's ability to utilise tuples from the support set as the number of videos increases. 

To analyse how many support videos are used, we train TRX with pairs ($\Omega{=}\{2\}$) and quadruples ($\Omega{=}\{4\}$) on SSv2 and Kinetics.  
{For each query tuple, we find the support set tuple with the maximum attention value.  We then count the number of support set videos per class which contain at least one maximal match, and average over all test tasks.}
Figure~\ref{fig:observed} presents the results for true and false, positive and negative, results.
The figure demonstrates that TRX successfully matches the query to tuples from multiple videos in the support set.
Most queries ($>$ 50\%) match to 2-3 videos in the support set.
Very few queries match to all~5~videos in the support set, particularly for higher cardinality tuples.
A similar distribution is seen for both datasets, however for SSv2, more true positive queries are matched to a single video in the support set.

\begin{SCfigure}[0.9]
    \includegraphics[scale=0.41,trim={0 20 15 10}]{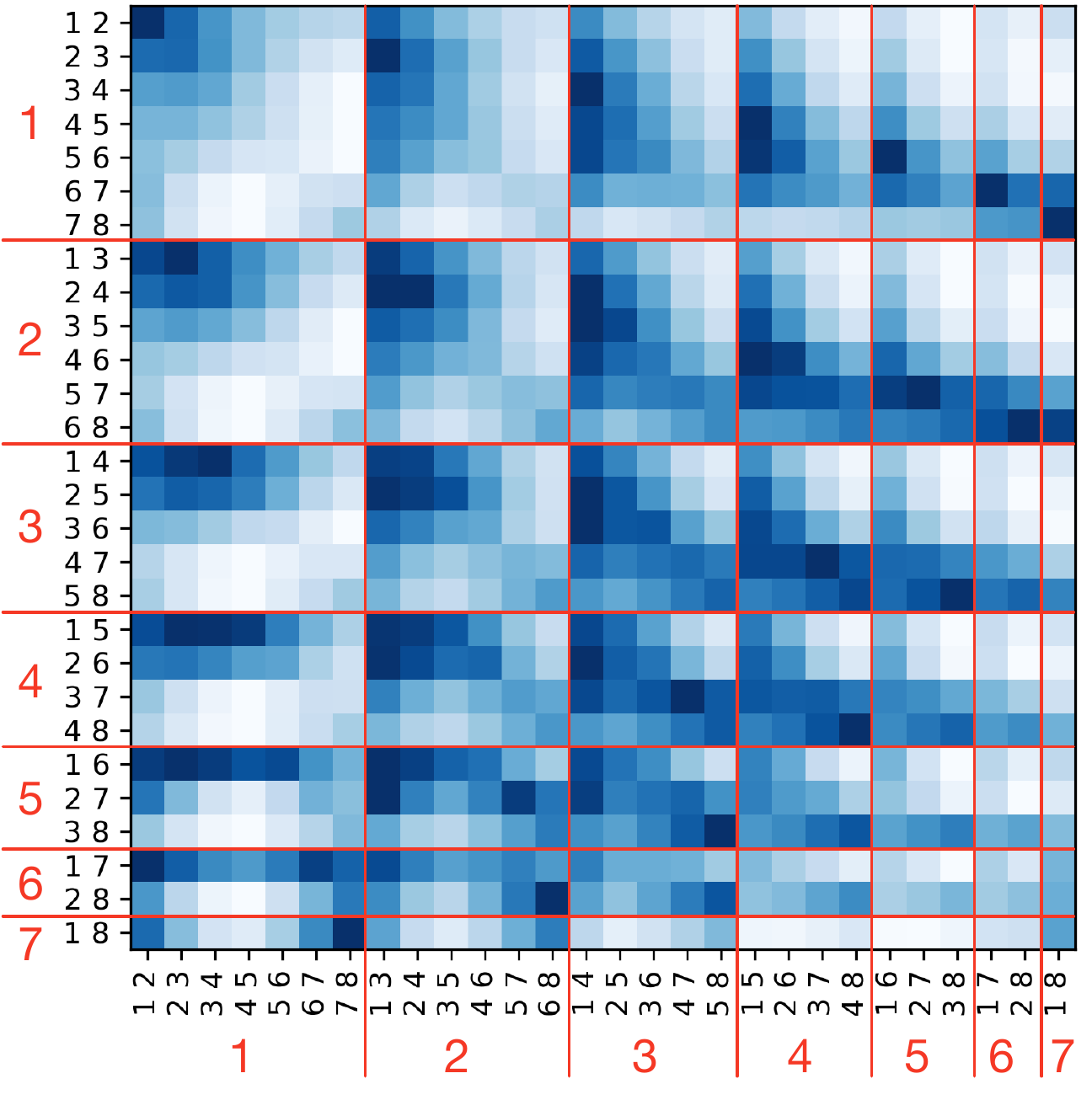}
  \caption{Summed attention over the SSv2$^\dagger$ test set, showing how query pairs (rows) match to support pairs (columns) from the TRX $\Omega{=}\{2,3\}$.  Numbers in red show the distance between the frames in the pair. \label{fig:confusion}
}
\end{SCfigure}

\vspace*{-2pt}
\subsubsection{Visualising tuple matches}\label{sec:ablation_look_pairs}
\vspace*{-5pt}

In addition to matching multiple support set videos, we visualise the tuple matches between the queries and the support set. 
Given a query tuple (row) and a support set tuple (col), we sum the attention values over the test set, and then normalise per row.
Fig.~\ref{fig:confusion} shows the summed attention values between all sets of pairs. 
While query pairs match frequently to corresponding support set pairs (i.e. same frame positions) in the support set, pairs are also matched to shifted locations (e.g. [1,2] with [2,4]) as well as significantly different frame distances (e.g. [6,7] with [1,7]). 

\vspace*{-10pt}

 \subsubsection{Varying the number of frames}\label{sec:ablation_runtime}
 
\begin{figure}[t]
\centering 
\includegraphics[width=\linewidth, trim={15 25 60 0}]{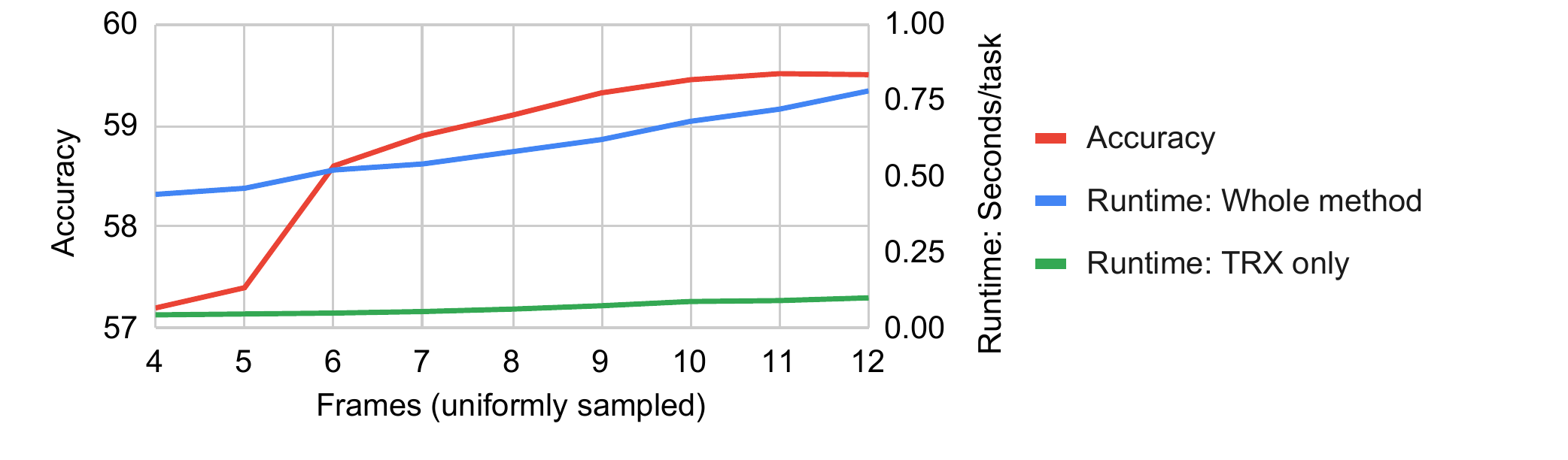}
\caption{SSv2$^\dagger$ accuracy (left y-axis) vs runtime analysis~(right y-axis in seconds/task) for TRX $\Omega=\{2,3\}$ as the number of sampled frames varies from 4 to 12 frames.}
\label{fig:frames}
\end{figure}

All previous results sample 8 frames from each video, in the query and support set.
This
allows us to directly compare to previous few-shot works that all consider 8 uniformly sampled frames~\cite{Zhu2020,Bishay2019,Cao2020}. 
To demonstrate TRX is scalable, we plot $\Omega=\{2,3\}$ results on SSv2 for the 5-way 5-shot task, sampling different numbers of frames~(4-12), and compare the accuracy and runtime in Fig.~\ref{fig:frames}.
Accuracy is comparable for $\ge 6$ frames. Importantly, the method's runtime scales linearly with the number of frames.
The TRX component only contributes a margin of the runtime and memory requirements of the network, with the ResNet-50 backbone dominating both needs.

\vspace*{-8pt}
\subsubsection{Random tuples in TRX}\label{sec:ablation_exhaustive}
\vspace*{-5pt}

\begin{figure}[t]
\centering
\subfloat[$\Omega{=}\{2\}$.\label{fig:rand2}]{\includegraphics[scale=0.38,trim={0 0 0 10}]{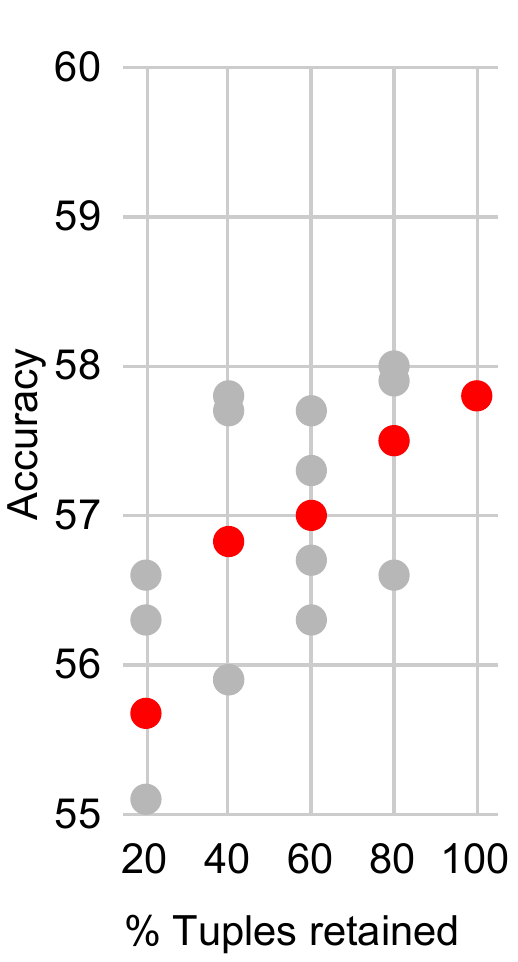}}\hspace{0mm}
\subfloat[$\Omega{=}\{3\}$.\label{fig:rand3}]{\includegraphics[scale=0.38,trim={0 0 0 10}]{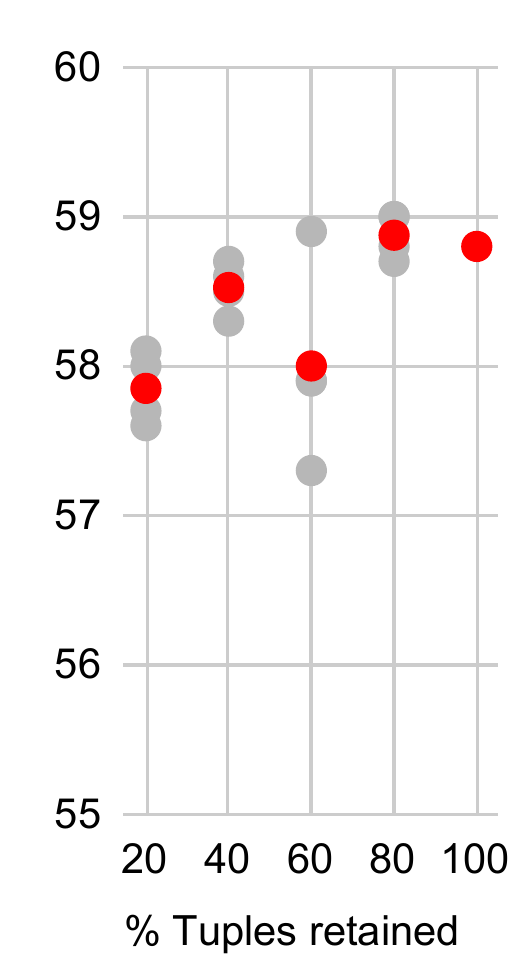}}\hspace{0mm}
\subfloat[$\Omega{=}\{4\}$.\label{fig:rand4}]{\includegraphics[scale=0.38,trim={0 0 0 10}]{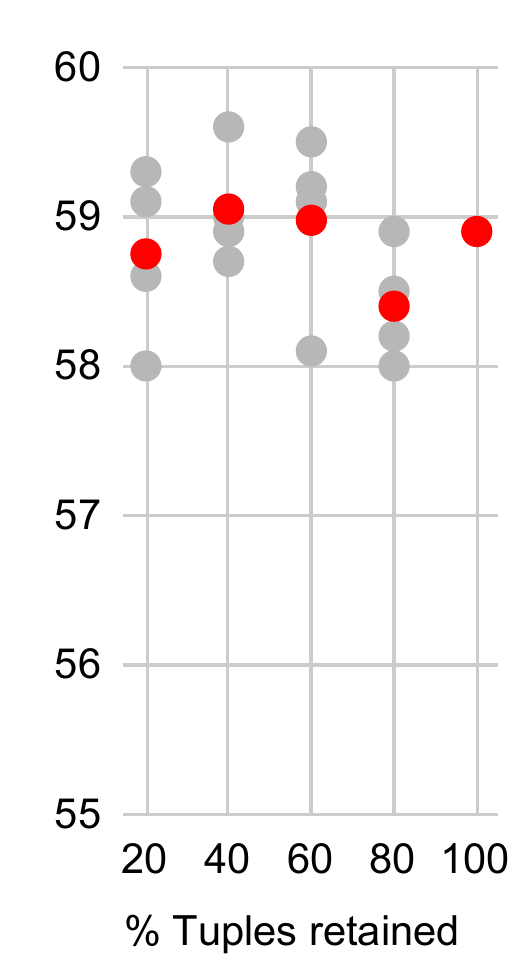}}\hspace{0mm}
\subfloat[$\Omega{=}\{2,3\}$.\label{fig:rand23}]{\includegraphics[scale=0.38,trim={0 0 0 10}]{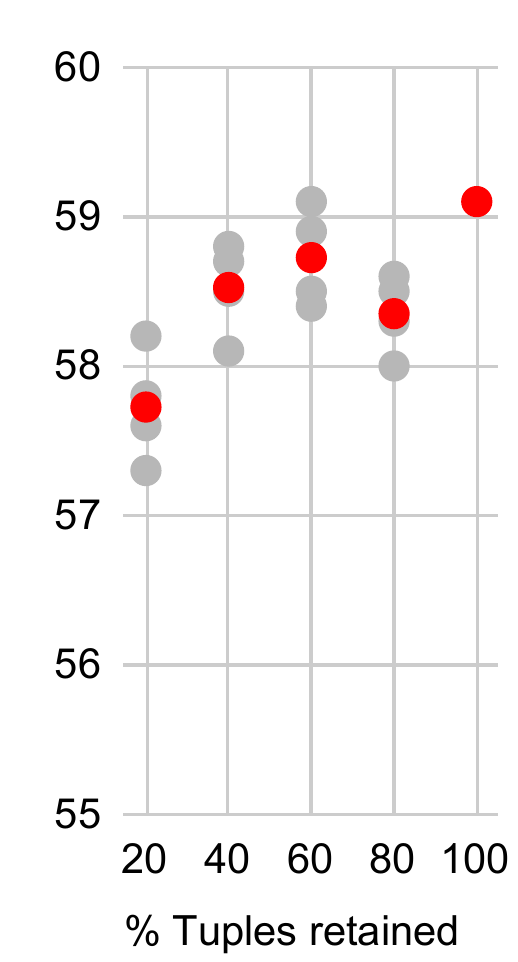}}
\vspace{-5pt}
\caption{Effect of retaining \% of tuples, selected randomly, on TRX, reported for SSv2$^\dagger$.  Grey dots indicate results of different runs, with averages in red.\vspace{-5pt}}
\label{fig:drop_tuples}
\end{figure}

All the above experiments have used exhaustive sets of tuples, e.g. every possible pair $(n_1,n_2)$ such that $1 \leq n_1 < n_2 \leq F$ for $\omega=2$.  To explore the impact of randomly sampling tuples, we experiment with 20, 40, 60 and 80\% of tuples retained for $\Omega{=}\{2\}$, $\{3\}$ and $\{4\}$, as well as a combined $\Omega{=}\{2,3\}$. We report four runs for each percentage, each with a different random selection of tuples.
\begin{table}[t]
\small
\begin{tabular}{lccccc}
\toprule 
&   \multicolumn{5}{c}{\% tuples retained} \\
\cmidrule(){2-6}
Cardinalities  \hspace{3mm}         &   20   &   40   &   60     &   80     &   100   \\ \midrule
$\Omega{=}\{2\}$        &   128 &   204 &   256 &   346 &   462\\ 
$\Omega{=}\{3\}$        &   228 &   390 &   540 &   702 &   844\\ 
$\Omega{=}\{4\}$        &   294 &   546 &   754 &   922 &   1152\\ 
$\Omega{=}\{2,3\}$      &   356 &   594 &   796 &   1048 &   1306\\
\bottomrule
\end{tabular}
\vspace{-5pt}
\caption{GPU usage in MiB when randomly dropping tuples, corresponding to experiments in Fig.~\ref{fig:drop_tuples}.\vspace{-10pt}}
\label{tab:drop_tuples}
\end{table}

Fig.~\ref{fig:drop_tuples}  shows that while retaining all tuples gives the best performance, some of the runs produce results comparable to exhaustive tuple selections for $\Omega{=}\{2,3\}$ and even outperform these for $\Omega{=}\{4\}$. The performance degrades quicker for $\Omega{=}\{2\}$.  The associated Tab.~\ref{tab:drop_tuples} compares the corresponding GPU usage. This shows it is possible to utilise fewer resources with comparable performance. A method for selecting tuples that maintain performance is left for future work.

\vspace{-5pt}
\section{Conclusion}
\vspace{-2pt}

This paper introduced Temporal-Relational CrossTransformers (TRX) for few-shot action recognition.  TRX constructs query-specific class prototypes by comparing the query to sub-sequences of all support set videos.  To model temporal relationships, videos are represented by ordered tuples of frames, which allows sub-sequences of actions at different speeds and temporal offsets to be compared.
TRX achieves state-of-the-art results on the few-shot versions of four datasets: Kinetics, Something-Something V2, HMDB51 and UCF101.
An extensive set of ablations shows how TRX observes multiple support set videos, the importance of tuple representations over single-frame comparisons, and the benefits of exploiting tuples of different cardinalities.
As future work, we aim to explore spatio-temporal versions of TRX. 

\vspace{5pt}
{
\small
\noindent \textbf{Acknowledgements} Publicly-available datasets were used for this work. 
This work was performed under the SPHERE Next Steps EPSRC Project EP/R005273/1.
Damen is supported by EPSRC Fellowship UMPIRE~(EP/T004991/1).
}
{\small
\bibliographystyle{ieee_fullname}
\bibliography{main}

\begin{thebibliography}{10}\itemsep=-1pt

\bibitem{Ba2016}
Jimmy~Lei Ba, Jamie~Ryan Kiros, and Geoffrey~E Hinton.
\newblock {Layer Normalization}.
\newblock {\em arXiv}, 2016.

\bibitem{Bateni2020}
Peyman Bateni, Raghav Goyal, Vaden Masrani, Frank Wood, and Leonid Sigal.
\newblock {Improved Few-Shot Visual Classification}.
\newblock In {\em Computer Vision and Pattern Recognition}, 2020.

\bibitem{Bishay2019}
Mina Bishay, Georgios Zoumpourlis, and Ioannis Patras.
\newblock {TARN: Temporal Attentive Relation Network for Few-Shot and Zero-Shot
  Action Recognition}.
\newblock In {\em British Machine Vision Conference}, 2019.

\bibitem{Cao2020}
Kaidi Cao, Jingwei Ji, Zhangjie Cao, Chien-Yi Chang, and Juan~Carlos Niebles.
\newblock {Few-Shot Video Classification via Temporal Alignment}.
\newblock In {\em Computer Vision and Pattern Recognition}, 2020.

\bibitem{Carreira}
Joao Carreira and Andrew Zisserman.
\newblock {Quo Vadis, Action Recognition? A New Model and the Kinetics
  Dataset}.
\newblock {\em Computer Vision and Pattern Recognition}, 2017.

\bibitem{Damen2018EPICKITCHENS}
Dima Damen, Hazel Doughty, Giovanni~Maria Farinella, Sanja Fidler, Antonino
  Furnari, Evangelos Kazakos, Davide Moltisanti, Jonathan Munro, Toby Perrett,
  Will Price, and Michael Wray.
\newblock {Scaling Egocentric Vision: The EPIC-KITCHENS Dataset}.
\newblock In {\em European Conference on Computer Vision}, 2018.

\bibitem{Deng2009}
Jia Deng, Wei Dong, Richard Socher, Li-Jia Li, Kai Li, and Li Fei-Fei.
\newblock {ImageNet: A Large-Scale Hierarchical Image Database}.
\newblock In {\em Computer Vision and Pattern Recognition}, 2009.

\bibitem{Doersch2020}
Carl Doersch, Ankush Gupta, and Andrew Zisserman.
\newblock {CrossTransformers: Spatially-Aware Few-Shot Transfer}.
\newblock In {\em Advances in Nerual Information Processing Systems}, 2020.

\bibitem{Dwivedi2019}
Sai~Kumar Dwivedi, Vikram Gupta, Rahul Mitra, Shuaib Ahmed, and Arjun Jain.
\newblock {ProtoGAN: Towards Few Shot Learning for Action Recognition}.
\newblock In {\em Computer Vision and Pattern Recognition}, 2019.

\bibitem{Feichtenhofer2019}
Christoph Feichtenhofer, Haoqi Fan, Jitendra Malik, and Kaiming He.
\newblock {Slowfast Networks for Video Recognition}.
\newblock In {\em International Conference on Computer Vision}, 2019.

\bibitem{Finn2017}
Chelsea Finn, Pieter Abbeel, and Sergey Levine.
\newblock {Model-Agnostic Meta-Mearning for Fast Adaptation of Deep Networks}.
\newblock In {\em International Conference on Machine Learning}, 2017.

\bibitem{Goyal2017}
Raghav Goyal, Vincent Michalski, Joanna Materzy, Susanne Westphal, Heuna Kim,
  Valentin Haenel, Peter Yianilos, Moritz Mueller-freitag, Florian Hoppe,
  Christian Thurau, Ingo Bax, and Roland Memisevic.
\newblock {The “Something Something” Video Database for Learning and
  Evaluating Visual Common Sense}.
\newblock In {\em International Conference on Computer Vision}, 2017.

\bibitem{He2016}
Kaiming He, Xiangyu Zhang, Shaoqing Ren, and Jian Sun.
\newblock {Deep Residual Learning for Image Recognition}.
\newblock In {\em Computer Vision and Pattern Recognition}, 2016.

\bibitem{Jiang2019}
Boyuan Jiang, Meng~Meng Wang, Weihao Gan, Wei Wu, and Junjie Yan.
\newblock {STM: SpatioTemporal and Motion Encoding for Action Recognition}.
\newblock In {\em International Conference on Computer Vision}, 2019.

\bibitem{Kang2019}
Bingyi Kang, Zhuang Liu, Xin Wang, Fisher Yu, Jiashi Feng, and Trevor Darrell.
\newblock {Few-Shot Object Detection via Feature Reweighting}.
\newblock In {\em International Conference on Computer Vision}, 2019.

\bibitem{Kuehnea}
H Kuehne, T Serre, H Jhuang, E Garrote, T Poggio, and T Serre.
\newblock {HMDB: A large video database for human motion recognition}.
\newblock In {\em International Conference on Computer Vision}, nov 2011.

\bibitem{Nichol}
Alex Nichol, Joshua Achiam, and John Schulman.
\newblock {On First-Order Meta-Learning Algorithms}.
\newblock {\em arXiv}, 2018.

\bibitem{Price}
Will Price and Dima Damen.
\newblock {Retro-Actions : Learning ‘Close' by Time-Reversing ‘Open'
  Videos}.
\newblock In {\em International Conference on Computer Vision Workshops}, 2019.

\bibitem{Requeima2019}
James Requeima, Jonathan Gordon, John Bronskill, Sebastian Nowozin, and
  Richard~E. Turner.
\newblock {Fast and Flexible Multi-Task Classification Using Conditional Neural
  Adaptive Processes}.
\newblock In {\em Advances in Nerual Information Processing Systems}, 2019.

\bibitem{Snell2017}
Jake Snell, Kevin Swersky, and Richard Zemel.
\newblock {Prototypical Networks for Few-Shot Learning}.
\newblock {\em Advances in Neural Information Processing Systems}, 2017.

\bibitem{Soomro2012}
Khurram Soomro, Amir~Roshan Zamir, and Mubarak Shah.
\newblock {UCF101: A Dataset of 101 Human Actions Classes From Videos in The
  Wild}.
\newblock {\em arXiv}, 2012.

\bibitem{triantafillou2019metadataset}
Eleni Triantafillou, Tyler Zhu, Vincent Dumoulin, Pascal Lamblin, Utku Evci,
  Kelvin Xu, Ross Goroshin, Carles Gelada, Kevin Swersky, Pierre-Antoine
  Manzagol, and Hugo Larochelle.
\newblock {Meta-Dataset: A Dataset of Datasets for Learning to Learn from Few
  Examples}.
\newblock In {\em International Conference on Learning Representations}, 2019.

\bibitem{Vaswani2017}
Ashish Vaswani, Noam Shazeer, Niki Parmar, Jakob Uszkoreit, Llion Jones,
  Aidan~N. Gomez, {\L}ukasz Kaiser, and Illia Polosukhin.
\newblock {Attention is all you need}.
\newblock In {\em Advances in Neural Information Processing Systems}, 2017.

\bibitem{Vinyals2016}
Oriol Vinyals, Charles Blundell, Timothy Lillicrap, Koray Kavukcuoglu, and Daan
  Wierstra.
\newblock {Matching Networks for One Shot Learning}.
\newblock In {\em Advances in Neural Information Processing Systems}, 2016.

\bibitem{Wanga}
Limin Wang, Yuanjun Xiong, Zhe Wang, Yu Qiao, Dahua Lin, Xiaoou Tang, and Luc
  {Van Gool}.
\newblock {Temporal Segment Networks: Towards Good Practices for Deep Action
  Recognition}.
\newblock In {\em European Conference on Computer Vision}, 2016.

\bibitem{Xu2017}
Zhongwen Xu, Linchao Zhu, and Yi Yang.
\newblock {Few-Shot Object Recognition from Machine-Labeled Web Images}.
\newblock In {\em Computer Vision and Pattern Recognition}, 2017.

\bibitem{Zhang2020}
Hongguang Zhang, Li Zhang, Xiaojuan Qi, Hongdong Li, Philip H~S Torr, and Piotr
  Koniusz.
\newblock {Few-shot Action Recognition with Permutation-invariant Attention}.
\newblock In {\em European Conference on Computer Vision}, 2020.

\bibitem{Zhang2018}
Ruixiang Zhang, Tong Che, Yoshua Bengio, Zoubin Ghahramani, and Yangqiu Song.
\newblock {Metagan: An Adversarial Approach to Few-Shot Learning}.
\newblock {\em Advances in Neural Information Processing Systems}, 2018.

\bibitem{Zhou2018}
Bolei Zhou, Alex Andonian, Aude Oliva, and Antonio Torralba.
\newblock {Temporal Relational Reasoning in Videos}.
\newblock In {\em European Conference on Computer Vision}, 2018.

\bibitem{Zhou_2018_ECCV}
Bolei Zhou, Alex Andonian, Aude Oliva, and Antonio Torralba.
\newblock {Temporal Relational Reasoning in Videos}.
\newblock In {\em European Conference on Computer Vision}, 2018.

\bibitem{Zhu2018}
Linchao Zhu and Yi Yang.
\newblock {Compound Memory Networks for Few-Shot Video Classification}.
\newblock In {\em European Conference on Computer Vision}, 2018.

\bibitem{Zhu2020}
Linchao Zhu and Yi Yang.
\newblock {Label Independent Memory for Semi-Supervised Few-shot Video
  Classification}.
\newblock {\em Transactions on Pattern Analysis and Machine Intelligence},
  14(8), 2020.

\bibitem{Zhu2021}
Xiatian Zhu, Antoine Toisoul, Juan-Manuel P{\'{e}}rez-R{\'{u}}a, Li Zhang,
  Brais Martinez, and Tao Xiang.
\newblock {Few-shot Action Recognition with Prototype-centered Attentive
  Learning}.
\newblock {\em arXiv}, 2021.

\end{thebibliography}
}

\newpage

\appendix

\section*{Appendix}

\section{X-Shot results}

\begin{table*}
\small
\begin{tabular}{llccccc}
\toprule 
&    & \multicolumn{5}{c}{Shot} \\
\cmidrule(){3-7}
Dataset \hspace{3mm} &   Method  \hspace{15mm}    &  \hspace{3mm} 1 \hspace{3mm}  &  \hspace{3mm} 2 \hspace{3mm}  &  \hspace{3mm} 3 \hspace{3mm}    &  \hspace{3mm} 4 \hspace{3mm}    &  \hspace{3mm} 5 \hspace{3mm}   \\ 
\midrule
           &    CMN \cite{Zhu2018}       & 60.5  & -  & -  & -  & 78.9  \\ 
           &    CMN-J \cite{Zhu2020}      &  60.5   &    70.0   &    75.6   &    77.3   &    78.9   \\
           &    TARN \cite{Bishay2019}       & 64.8  & -  & -  & -  & 78.5  \\ 
Kinetics   &    ARN \cite{Zhang2020}       & 63.7  & -  & -  & -  & 82.4  \\ 
           &    OTAM \cite{Cao2020}       & \textbf{73.0}  & -  & -  & -  & 85.8  \\ 
           &    Ours - TRX $\Omega{=}\{1\}$       & 63.6  & 75.4  & 80.1  & 82.4  & 85.2  \\ 
           &    Ours - TRX $\Omega{=}\{2,3\}$       & 63.6  & \textbf{76.2}  & \textbf{81.8}  & \textbf{83.4}  & \textbf{85.9}  \\ 
\midrule
                    &     CMN-J \cite{Zhu2020}     &  \bf{36.2}   &    42.1   &    44.6   &    47.0   &    48.8   \\
SSv2$^*$ &    Ours - TRX $\Omega{=}\{1\}$       &    34.9  & 43.4   &    47.6   &    50.9   &    53.3   \\ 
                    &    Ours - TRX $\Omega{=}\{2,3\}$       &  36.0   &  \textbf{46.0}   &  \textbf{51.9}   &    \textbf{54.9}   &    \textbf{59.1}   \\
\midrule
                    &    OTAM \cite{Cao2020}       & \textbf{42.8}  & -  & -  & -  & 52.3  \\ 
SSv2$^\dagger$ &    Ours - TRX $\Omega{=}\{1\}$       & 38.8     & 49.7   & 54.4     & 58.0      & 60.6      \\ 
                    &    Ours - TRX $\Omega{=}\{2,3\}$       &  42.0   &  \textbf{53.1}   &  \textbf{57.6}   &    \textbf{61.1}   &    \textbf{64.6}   \\
\bottomrule
\end{tabular}
\caption{Comparison to few-shot video works on Kinetics (split from \cite{Zhu2020}) and Something-Something V2 (SSv2) ($^\dagger$: split from \cite{Zhu2020}  $^*$: split from~\cite{Cao2020}). Results are reported as the shot, \ie number of support set videos per class, increases from 1 to 5.  
-: Results not available in published works.
}
\label{tab:all_shots}
\end{table*}

In the main paper, we introduced Temporal-Realational CrossTransformers (TRX) for few-shot action recognition.  They are designed specifically for $K$-shot problems where $K>1$, as TRX is able to match sub-sequences from the query against sub-sequences from multiple support set videos.

Table \ref{tab:main_results} in the main paper shows results on the standard 5-way 5-shot benchmarks on Kinetics \cite{Carreira}, Something-Something V2 (SSv2) \cite{Goyal2017}, HMDB51 \cite{Kuehnea} and UCF101 \cite{Soomro2012}.
For completeness we also provide 1-, 2-, 3-, 4- and 5-shot results for TRX with $\Omega{=}\{1\}$ (\ie frame-to-frame comparisons) and $\Omega{=}\{2,3\}$ (\ie pair and triplet comparisons) on the large-scale datasets Kinetics and SSv2.  These are in Table~\ref{tab:all_shots} in this appendix, where we also list results from all other works which provide these scores.

For 1-shot, in Kinetics, TRX performs similarly to recent few-shot action-recognition methods \cite{Zhu2018,Bishay2019,Zhang2020}, but these are all outperformed by OTAM \cite{Cao2020}.  OTAM works by finding a strict alignment between the query and single support set video per class.  It does not scale as well as TRX when $K>1$, shown by TRX performing better on the 5-shot benchmark.
This is because TRX is able to match query sub-sequences against similar sub-sequences in the support set, and importantly ignore sub-sequences (or whole videos) which are not as useful.  
Compared to the strict alignment in OTAM~\cite{Cao2020}, where the full video is considered in the alignment, TRX can exploit several sub-sequences from the same video, ignoring any distractors.  Despite not being as well suited to 1-shot problems, on SSv2 TRX performs similarly to OTAM.  2-shot TRX even outperforms 5-shot OTAM.
Table~\ref{tab:all_shots} again highlights the importance of tuples, shown in the main paper, where TRX with $\Omega{=}\{2,3\}$ consistently outperforms $\Omega{=}\{1\}$.

\begin{figure}
\centering
\includegraphics[scale=0.5, trim={0 10 20 5}]{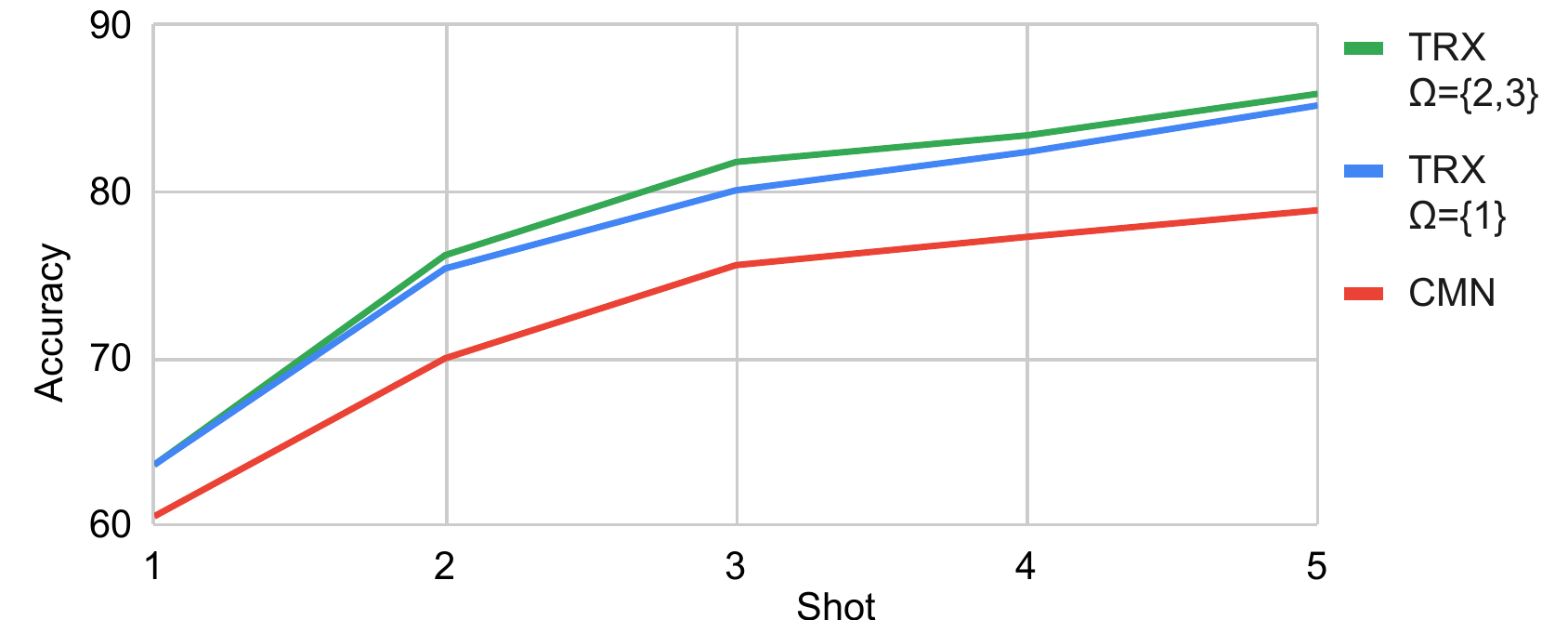}
\caption{Comparing CMN~\cite{Zhu2020} results to TRX for X-shot 5-way, for $1 \le X \le 5$ on Kinetics. TRX benefits from increasing the number of of videos in the support set, both for $\Omega{=}\{1\}$ and $\Omega{=}\{2,3\}$.}
\label{fig:kshotkinetics}
\end{figure}

Figure 5 in the main paper shows how TRX scales on SSv2 compared to CMN \cite{Zhu2018,Zhu2020}, which also provides X-shot results ($1 \le X \le 5)$.  The equivalent graph for Kinetics is shown in Fig. \ref{fig:kshotkinetics} here. This confirms TRX scales better as the shot increases.  There is less of a difference between TRX with $\Omega{=}\{1\}$ and $\Omega{=}\{2,3\}$, as Kinetics requires less temporal knowledge to discriminate between the classes than SSv2 (ablated in Sec. 4.3.1 and 4.3.2 in the main paper).

\section{The impact of positional encoding}\label{sec:ablation_pe}

\begin{table}
\small
\centering 
\begin{tabular}{lccc}
\toprule
Method              & Positional Encoding        &   Kinetics    &   SSv2$^\dagger$ \\ \midrule
$\Omega{=}\{1\}$    &  $\times$                 & 85.2          & 53.0       \\
$\Omega{=}\{1\}$    & \checkmark       &    85.2       & 53.3       \\
$\Omega{=}\{2,3\}$  &   $\times$                & 85.5          &  58.5      \\
$\Omega{=}\{2,3\}$  &  \checkmark       &   {\bf 85.9}        & {\bf 59.1}       \\
\bottomrule
\end{tabular}
\vspace{-5pt}
\caption{The importance of incorporating positional encoding for single frames and the proposed model ${\Omega{=}\{2,3\}}$.}
\label{tab:positional_encoding}
\end{table}

TRX adds positional encodings to the individual frame representations before concatenating them into tuples.  Table~\ref{tab:positional_encoding} shows that adding positional encodings improves SSv2 for both single frames and higher-order tuples (by +0.3\% and +0.6\% respectively).
For Kinetics, performance stays the same as single frames and improves slightly with tuples (+0.4\%) for the proposed model.
Overall, positional encoding improves the results marginally for TRX.

\end{document}